\makeatletter\@ifundefined{ifanonymous}{\newif\ifanonymous\anonymousfalse}{}\makeatother
\documentclass[11pt,a4paper]{article}
\usepackage{labarxiv}
\labauthors{\au{Haomin Luo}{1,2}, \au{Ersha Li}{2,3}}
\labaffils{\af{1}{University of Cambridge}, \af{2}{Models$^2$ AI Lab}, \af{3}{The Hong Kong University of Science and Technology}}
\labitems{%
  \lablinkitem{\faGithub}{https://github.com/hl682}{github.com/hl682}\labsep
  \labmailitem{hl682@cam.ac.uk}\labsep
  \labwebitem{https://www.models2.com}{models2.com}}
\labfootnote{\noindent\itshape Corresponding author: \href{mailto:hl682@cam.ac.uk}{\textcolor{labblue}{hl682@cam.ac.uk}}\\
  \noindent\upshape\textcopyright{} 2026 Models$^2$. All rights reserved.}
\labacknowledgments{%
  This research is funded by Models$^2$ AI and receives funding from the
  Shanghai Municipal \mbox{Government}, the Jiaxing Municipal Government, the Xuhui
  District Government of Shanghai, and the Xiuzhou District Government of Jiaxing.
  The research is supported by the Model Speed Space Large Model Innovation
  Ecosystem Community and the Jiarongshe Artificial Intelligence Incubation Programme.}

\newtheorem{proposition}{Proposition}
\newtheorem{corollary}{Corollary}
\newcounter{algorithmctr}

\labtitle{Escaping the One-Step Trap: Temporal Shortcuts and Stopped-Option Semigroups for Jumpy World Models}

\lababstract{%
Standard world models simulate the future one primitive action at a time. Under a fixed test-time compute budget, this \emph{one-step trap} spends every neural evaluation on short-term physics rather than long-horizon decisions. \emph{Jumpy world models} escape this bottleneck by predicting the multi-step outcome of executing an option or macro-skill in $O(1)$ evaluations—unifying few-step generative shortcuts along denoising time with option-semigroup composition across physical time. We formalize this synthesis as a five-tuple \emph{stopped-option semigroup} that separates internal skill stopping from episode termination, conditions outcome readouts jointly on the realized endpoint, and composes variable-length jumps algebraically ($h + h \to 2h$). Across 319 trained models and over 770,000 closed-loop planning episodes spanning eight model families, four discrete and continuous control tasks, and five latent geometries, we establish four principles for jumpy planning. First, temporal shortcuts amplify subgoal reachability rather than raw jump duration: planning fails on non-covering controllers (0\% on \texttt{PinPad} and \texttt{Sokoban}) and gains over 41 percentage points as subgoal skills unlock, even when option durations are held constant at $\sim\!56$ steps. Second, where a jump lands dictates how to discount it: predicting duration and return marginally without conditioning on the realized endpoint blurs fast subgoal arrivals with slow timeouts, inflating option-Bellman backup error up to $23.5\times$ and losing 29.3 percentage points on stochastic risky-shortcut tasks. Third, bounded latent norms still hide a bootstrap manifold trap: on convex simplices, two-step flow endpoints drift into an unsupervised interior where dyadic bootstrapping amplifies return drift to $4.85\times 10^8$; anchoring readouts on predicted endpoints and transporting latents along constant-norm spherical geodesics eliminate divergence across all 27/27 anchored full-budget runs. Fourth, deterministic expectation dominates standard control while spherical semigroup flows unlock extrapolation and multimodality: with full-span data under near-deterministic dynamics, a single-step endpoint-anchored expectation model reaches 54.75\% solve rate at just 256 dynamics evaluations per decision—surpassing multi-step flow baselines given $16\times$ more compute—while committing to chosen options cuts episode compute by another $25.5\text{--}27.2\times$. Under truncated offline spans ($h \le 16$) or bimodal stochastic transitions, anchored spherical semigroup flows achieve the lowest $8\times$ horizon-extrapolation error (0.193) and gain $+10.2$ percentage points from multi-particle planning.}

\labkeywords{world models, temporal abstraction, options, test-time planning, shortcut flow models, Riemannian manifolds}

\begin{document}
\makelabtitle

\section{Introduction}

Human planning rarely simulates muscle twitches millisecond by millisecond; to cross a building, we reason in leaps from doorway to doorway. Most learned world models do the opposite \citep{hansen2024tdmpc2,hafner2025mastering}. Because a primitive dynamics model advances imagination by a single environment tick per neural function evaluation (NFE), a planner operating under a fixed test-time budget of $B$ evaluations per candidate sequence can look ahead at most $B$ primitive actions \citep{wan2019planning,zhang2026knowing}. On tasks where meaningful consequences unfold over tens or hundreds of steps, search burns its entire compute budget on local physics without ever seeing a subgoal—a pathology \citet{sutton1999between} termed the \emph{one-step trap}.

Temporally abstract or \emph{jumpy} world models escape this trap by predicting the state, elapsed time, and cumulative reward at the end of a multi-step option in $O(1)$ dynamics evaluations, stretching lookahead from $B$ primitive ticks to $B$ skill executions \citep{silver2012compositional,freed2023learning,rodriguezsanchez2024learning,farebrother2026compositional}. Recent breakthroughs in few-step generative transport—Shortcut Models \citep{frans2025one}, self-distilled Flow Maps \citep{boffi2025flow}, MeanFlow \citep{geng2025mean}, and Temporal Difference Flows \citep{farebrother2025temporal}—point to a compelling mathematical synthesis. Just as two half-steps along denoising time $\tau \in [0,1]$ distill into one fast generation jump ($\Delta_\tau$), two half-spans of an option along physical time $t$ can compose into a doubly long temporal jump ($\Delta_t: h + h \to 2h$). In principle, such a \emph{bi-axial shortcut world model} can leap across 32 to 128 environment steps in one or two forward passes without step-by-step rollout drift.

\begin{labfigure}[!t]{End-to-end architecture of five-tuple stopped-option semigroup world models (\textbf{Stage 1--Stage 3}). \textbf{(1)} Executing option $\omega$ under horizon cap $h$ yields a joint stopped-option five-tuple $(Z', D, R, m, e)$ separating internal skill termination $m$ from episode absorption $e$; conditioning scalar readouts jointly on $Z'$ eliminates both the discount--value covariance bias $-\mathrm{Cov}(\tilde{\Gamma}, V(Z'))$ and concave \texttt{Symlog} duration plug-in bias incurred by marginal heads. \textbf{(2)} Bi-axial self-distillation combines empirical flow-map matching with stop-gradient predicted-endpoint readout anchoring (75\% batches) and dyadic $h + h \to 2h$ semigroup composition with decision value-equivalence (25\% batches). \textbf{(3)} Fixed-budget chance-node receding-horizon MPC averages $N_p$ Monte Carlo particles per candidate option sequence prior to maximization and commits to the chosen option until termination.}{fig:arch}
\labincludegraphics{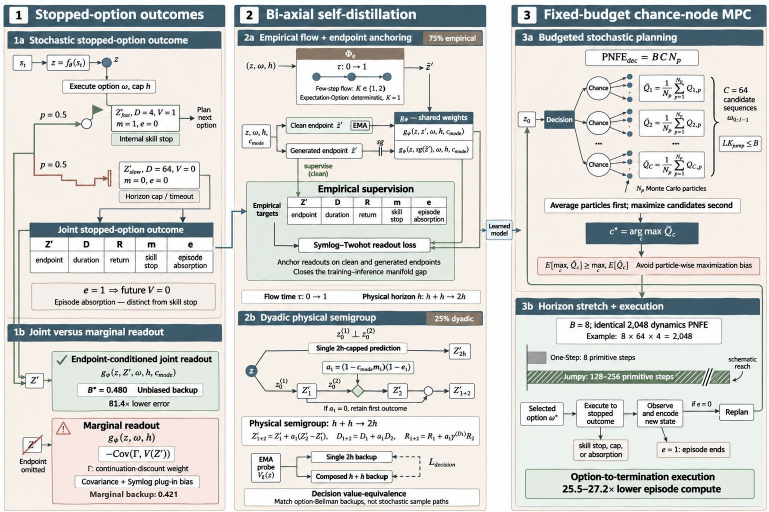}
\end{labfigure}

In practice, however, turning multi-step generative shortcuts into reliable closed-loop planning exposes three hidden traps that one-step world models never encounter—and that prior jumpy formulations entangle:
\begin{enumerate}
\item \textbf{The Bifurcation Discount Trap.} Primitive steps always take $\Delta t = 1$ tick, so discounting is fixed at $\gamma$. Options take \emph{random} durations $D$. When a skill can either reach a rewarding doorway quickly ($D=4$, high future value $V$) or wander until timeout ($D=64$, low $V$), predicting duration and reward \emph{marginally} from $(z,\omega,h)$ without conditioning on \emph{where the jump landed} ($Z'$) averages fast successes with slow dead-ends—systematically mispricing option values even when marginal predictions are exact.
\item \textbf{The Interior Manifold Trap.} Modern self-predictive world models avoid pixel reconstruction by bounding latent states on convex simplices (\texttt{SimNorm}; \citealp{lavoie2023simplicial,hansen2024tdmpc2}). Yet bounded norm does not imply shared support: two-step Euler flow trajectories cut across the interior of the simplex, landing in a higher-entropy region where clean encoder targets never supervise the reward head. When two-hop $\Delta_t$ bootstrapping queries this unanchored interior twice per horizon doubling, tiny positive errors self-amplify—even for $\gamma < 1$—until predicted returns saturate at $10^8$.
\item \textbf{The Determinism--Distribution Trade-off.} Multi-step generative flows pay extra solver evaluations per jump to model outcome distributions. Under a strict per-decision compute budget, when does distributional semigroup transport actually beat a single-step ($K=1$) deterministic expectation jump?
\end{enumerate}

We resolve these questions through a unified \textbf{five-tuple stopped-option semigroup} $(Z', D, R, m, e)$ paired with a large-scale controlled dissection across \textbf{319 trained world models and more than 770,000 closed-loop receding-horizon MPC episodes} (Figures~\ref{fig:arch} and~\ref{fig:coverage_frontier}). Our formulation explicitly separates internal skill stopping $m$ (where the planner resumes control at $Z'$ with future value $V(Z')$) from environment episode absorption $e$ (where future value vanishes), yielding exact algebraic $h + h \to 2h$ closure across variable-length jumps. Holding encoders, replay splits, categorical scalar codecs, and chance-node MPC budgets fixed across eight model families, four discrete and continuous benchmarks (\texttt{FourRooms}, \texttt{PinPad}, \texttt{Sokoban}, and \texttt{AntMaze}), and targeted causal interventions, we establish four core claims:

\begin{labfigure}[!t]{Behavioral subgoal prerequisite, compute frontier, and matched-lookahead control. \textbf{(a)} Within the \emph{same} $\mathcal{O}_{\mathrm{skill}}$ library at matched $\sim\!56$-step option durations, directional-only options (\texttt{L0}, 0 subgoals) fail on \texttt{PinPad} (0.0\%), whereas progressively unlocking $\{1,2,3,4\}$ subgoal skills (\texttt{L1--L4}) yields $+41.2$ to $+43.4$ percentage points paired four-task gains. \textbf{(b)} Across Dynamics Transition PNFE budgets from 256 to 4,096 per decision under $\mathcal{O}_{\mathrm{skill}}$ (plotting both $(B=4, N_p=4)$ filled markers and $(B=16, N_p=1)$ open markers at 1,024 PNFE, with across-seed SEM bands), single-step \texttt{Expectation-Option} Pareto-dominates all flow baselines, while chained short-span flows (\texttt{Denoise-Only}) are budget-infeasible at $B=4$. \textbf{(c)} Restricting \texttt{Expectation-Option} to $J=4$ jumps (nominal cap 128, matching two-step flows at half their 2,048 PNFE) reaches 55.21\% (\texttt{Val-best}), ruling out lookahead-depth confounding.}{fig:coverage_frontier}
\labincludegraphics{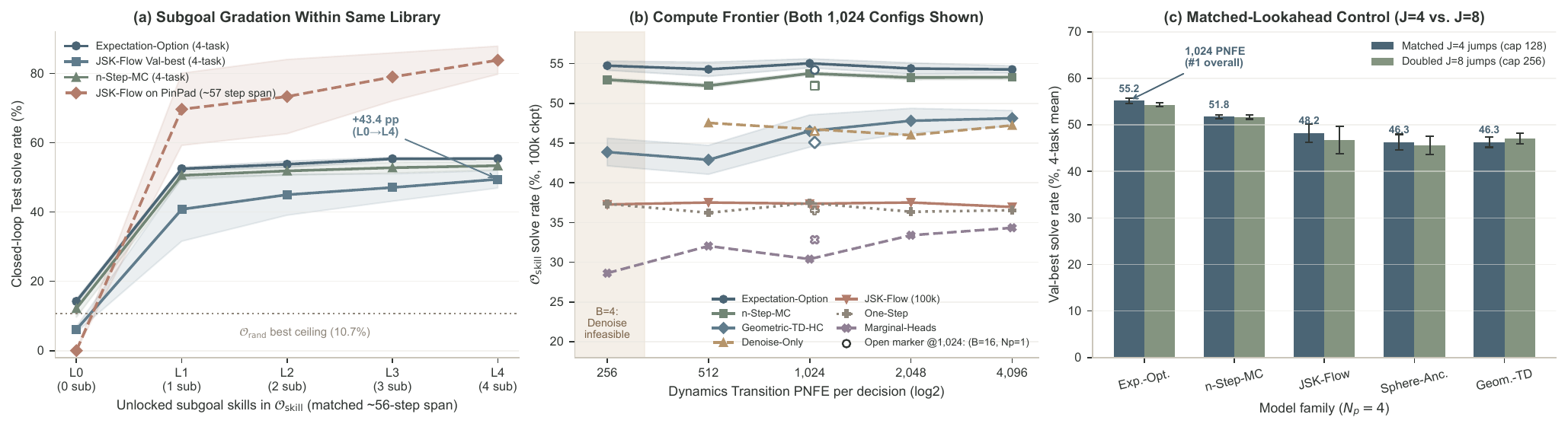}
\end{labfigure}

\begin{itemize}
\item \textbf{Temporal shortcuts multiply subgoal reachability, not raw jump length.} Jumpy planning cannot invent credit assignment across bottlenecks that options never visit: every architecture scores 0.00\% on \texttt{PinPad} and \texttt{Sokoban} under random controllers ($\mathcal{O}_{\mathrm{rand}}$), versus 89.75\%--90.08\% on \texttt{PinPad} with subgoal-covering skills ($\mathcal{O}_{\mathrm{skill}}$). Even \emph{within} $\mathcal{O}_{\mathrm{skill}}$ at matched $\sim\!56$-step option durations, directional-only options score 0.00\% on \texttt{PinPad}, whereas progressively unlocking $\{1,2,3,4\}$ subgoal skills yields monotonic $+41.21$ to $+43.38$ percentage points four-task gains (Figure~\ref{fig:coverage_frontier}a).
\item \textbf{Joint endpoint conditioning eliminates marginal covariance and \texttt{Symlog} plug-in bias.} We prove an exact three-term error identity (Proposition~\ref{prop:bifurcation} and Corollary~\ref{cor:symlog}) showing how marginal \texttt{Symlog-Twohot} readouts combine discount--value covariance $-\mathrm{Cov}(\tilde{\Gamma}, V(Z'))$ with concave geometric-mean duration underestimation, closing analytical and learned corridor backups to $5\times 10^{-5}$ ($81.4\times$ bias reduction). Across benchmarks, marginal heads inflate option-Bellman error by $3.8\text{--}23.5\times$, collapse \texttt{PinPad} MPC to 25.08\% (100k) / 38.08\% (\texttt{Val-best}), and trail anchored joint flows by 29.33 percentage points on bimodal risky-shortcut control.
\item \textbf{Readout anchoring and constant-norm spherical geodesics cure the semigroup manifold ratchet.} We prove that two-hop $\Delta_t$ composition amplifies unanchored interior return bias by $(1 + \hat{a}_1\gamma^{\hat{D}_1}) > 1$ per dyadic doubling (Proposition~\ref{prop:ratchet}), and isolate the $4.85\times 10^8$ blowup to the return head via zero-retraining single-head swaps ($0.129$ MAE). Supervising readouts directly on stop-gradient predicted endpoints (\texttt{JSK-Flow-Anchored}: 0/15 divergences on \texttt{SimNorm}) and transporting latents via Riemannian geodesics on $\mathbb{S}^{63}(8)$ (\texttt{JSK-Flow-Sphere-Anchored}: 0/12 divergences, $+20.69$ percentage points closed-loop lift over unanchored spherical flows) eliminate divergence across all \textbf{27/27} anchored 100k runs.
\item \textbf{A sharp regime law separates $K=1$ expectation models from spherical semigroup flows.} When offline trajectories cover full option spans under near-deterministic dynamics, $K=1$ \texttt{Expectation-Option} dominates every compute budget (Figure~\ref{fig:coverage_frontier}b,c)—reaching \textbf{54.75\%} at \textbf{256 dynamics evaluations per decision} (16.95 ms), beating both empirical flow maps (\texttt{n-Step-MC}, 53.29\%) and occupancy flows (\texttt{Geometric-TD-HC}, 48.12\%) at 4,096 evaluations ($16\times$ compression), while option-to-termination execution cuts episode evaluations by another \textbf{25.5--27.2$\times$} (over $400\times$ combined episode savings over step-by-step one-step planning). Conversely, when offline spans are truncated ($H_{\mathrm{train}} \le 16$) or transitions bifurcate stochastically, \texttt{JSK-Flow-Sphere-Anchored} achieves the lowest $8\times$ extrapolated open-loop error (0.193) and unlocks $+10.17$ percentage points from multi-particle ($N_p=4$) planning.
\end{itemize}

\section{Related Work}

\textbf{Options and Expectation World Models.} The options framework models temporal abstraction through initiation sets, internal controllers, and stochastic stopping rules \citep{sutton1999between}, enabling compositional option-Bellman backups \citep{silver2012compositional}. Because full successor distributions scale poorly to high-dimensional states, \citet{wan2019planning} and \citet{kudashkina2021planning} analyzed \emph{expectation models} that predict expected next features, characterizing when expectation backups remain value-equivalent and where nonlinear value curvature introduces bias. Subsequent work learned temporally abstract latent dynamics offline \citep{freed2023learning,jiralerspong2023forecaster,rodriguezsanchez2024learning}. Our study connects expectation theory to modern latent receding-horizon MPC: when scalar readouts are anchored directly on predicted endpoints $(z,\hat{z}',\omega,h)$, deterministic expectation models avoid few-step flow manifold drift and dominate near-deterministic control, while explicitly revealing their bias boundary under bimodal stochastic transitions.

\textbf{Generative Occupancy, Shortcut, and Riemannian Flow Maps.} Generative RL models replace mean regression with distributional transport, including Temporal Difference Flows \citep{farebrother2025temporal} and horizon-conditioned compositional occupancy flows (\texttt{Geometric-TD-HC}; \citealp{farebrother2026compositional}). Concurrently, generative modeling compressed ODE integration into one or two steps via flow-time ($\Delta_\tau$) self-distillation, spanning Shortcut Models \citep{frans2025one}, self-distilled Flow Maps \citep{boffi2025flow}, and MeanFlow \citep{geng2025mean}, recently generalized to Riemannian manifolds \citep{chen2024flow,davis2026generalised,woo2026riemannian}. In world modeling, however, $\Delta_\tau$ shortcuts only accelerate single-jump sampling; extrapolating physical horizon $h$ requires a second semigroup operator ($\Delta_t$) across environment time. We disentangle $\Delta_\tau$ from $\Delta_t$, proving how unanchored $\Delta_t$ composition triggers a $(1 + \hat{a}_1\gamma^{\hat{D}_1})$ readout ratchet on convex domains and how readout anchoring plus spherical geodesic transport resolve it.

\textbf{Bounded Latent Embeddings and Categorical Scalar Readouts.} Self-predictive world models without pixel reconstruction rely on bounded projections such as simplicial embeddings (\texttt{SimNorm}; \citealp{lavoie2023simplicial,hansen2024tdmpc2}), categorical two-hot cross-entropy with \texttt{symexp} decoding \citep{farebrother2024stop,hafner2025mastering,hafner2025training}, and latent value alignment \citep{grimm2020value,jiang2026boosting}. We show two previously uncharacterized interactions in this stack: first, point decoding via $\mathrm{symexp}(\mathbb{E}[\mathrm{symlog}(D)])$ yields a geometric-mean duration plug-in shift under marginal factorization; second, convex simplex bounds permit boundary-to-interior support separation under finite-step Euler flows.

\section{Stopped-Option Semigroups and Bi-Axial Shortcut World Models}
\label{sec:method}

We consider a Markov decision process $(\mathcal{S}, \mathcal{A}, P, r, \gamma)$ equipped with a library $\Omega$ of $|\Omega|=8$ Markov options $\omega \in \Omega$, each executing feedback policy $\pi_\omega(a \mid s)$ with per-step internal stopping indicator $m_t \in \{0,1\}$ distinct from environment episode absorption $e_t \in \{0,1\}$. An encoder $f_\theta: \mathcal{S} \to \mathcal{Z} \subset \mathbb{R}^d$ ($d=64$) maps observations to latent states alongside an exponential moving average (EMA) target encoder $f_{\bar{\theta}}$.

\subsection{Fixed-Budget Chance-Node Jumpy MPC and Exact Query Accounting}

At decision state $s_0$ with encoded state $z_0 = f_\theta(s_0)$ (computed once per decision), a chance-node receding-horizon planner evaluates $C=64$ candidate option sequences $\omega_{0:J-1}^{(c)} \in \Omega^J$ across $N_p$ independent Monte Carlo particles under a strict per-candidate dynamics budget $B$:
\begin{equation}
\mathrm{Dynamics\text{-}PNFE}_{\mathrm{dec}} = B \cdot C \cdot N_p, \qquad J \cdot K_{\mathrm{jump}} \le B,
\label{eq:pnfe}
\end{equation}
where $K_{\mathrm{jump}}$ is the number of neural dynamics evaluations per macro-jump of nominal cap $h_{\mathrm{plan}}$. For each candidate $c$, the planner averages cumulative discounted returns across all $N_p$ particles \emph{before} selecting $c^\star = \arg\max_c \frac{1}{N_p}\sum_{p=1}^{N_p} \hat{Q}_{c,p}$, avoiding the $\mathbb{E}[\max_c \hat{Q}_c] \ge \max_c \mathbb{E}[\hat{Q}_c]$ optimism bias of particle-wise maximization (Figure~\ref{fig:arch}, Stage~3a).

Distinguishing \emph{Dynamics Transition PNFE} from total network row queries makes per-decision cost transparent. At the default budget $B=8, C=64, N_p=4$ ($\mathrm{Dynamics\text{-}PNFE}_{\mathrm{dec}} = 2{,}048$, $256$ particle trajectories):
\begin{itemize}
\item \textbf{Two-step jumpy flows} ($K_{\mathrm{jump}}=2$, $J=4$ jumps of cap $h_{\mathrm{plan}}=32$, nominal lookahead $128$): execute $1 + 256\times(8\text{ dyn} + 4\text{ read} + 2\text{ probe}) = \mathbf{3{,}585}$ total row queries ($24.39\text{--}33.36$ ms/decision).
\item \textbf{Chained short-span flows} (\texttt{Denoise-Only}, trained on $h \le 8$ with $K=2$ per sub-jump): require $K_{\mathrm{jump}} = 4\times 2 = 8$ evaluations for a single $32$-step jump ($J=1$, $3{,}585$ queries) and are \emph{budget-infeasible} ($K_{\mathrm{jump}} > B$) at $B=4$.
\item \textbf{Single-step jumpy models} (\texttt{Expectation-Option}, \texttt{Time-Shortcut}; $K_{\mathrm{jump}}=1$, $J=8$ jumps, nominal lookahead $256$): execute $4$ additional readout evaluations across the doubled jump depth, totaling $1 + 256\times(8\text{ dyn} + 8\text{ read} + 2\text{ probe}) = \mathbf{4{,}609}$ row queries ($26.75$ ms/decision at $B=8, N_p=4$; only $2{,}305$ queries and $\mathbf{16.95}$ ms/decision at $B=4, N_p=1$).
\item \textbf{Primitive \texttt{One-Step}} ($K_{\mathrm{jump}}=1, h_{\mathrm{plan}}=1, J=8$, nominal lookahead $8$): additionally queries a behavior-cloned latent policy $\hat{\pi}(a \mid z,\omega)$ at every step, totaling $1 + 256\times(8\text{ dyn} + 8\text{ policy} + 8\text{ read} + 2\text{ probe}) = \mathbf{6{,}657}$ row queries ($31.81$ ms/decision).
\end{itemize}
After selecting $\omega_0^\star$, we evaluate both \emph{primitive replanning} (executing one primitive action $a_0 \sim \pi_{\omega_0^\star}(\cdot\mid s_0)$ and replanning every step) and \emph{option-to-termination execution} (executing $\omega_0^\star$ until skill stop, cap $h_{\mathrm{plan}}$, or episode absorption).

\subsection{The Five-Tuple Stopped-Option Semigroup}

Executing Markov option $\omega$ from environment state $s_t$ under horizon cap $h \in \mathcal{H} = \{1, 2, 4, 8,\allowbreak 16, 32, 64, 128\}$ and mode flag $c_{\mathrm{mode}} \in \{0,1\}$ ($c_{\mathrm{mode}}=1$ for stopped-option rollouts halting on $m \lor e$; $c_{\mathrm{mode}}=0$ for fixed-span rollouts halting only on $e$ or cap $h$) induces stopping index $\tau_{\mathrm{stop}} = \min\{k \in \{1,\dots,h\} : e_{t+k}=1 \lor (c_{\mathrm{mode}}=1 \land m_{t+k}=1)\}$ (or $h$ if empty) and five-tuple
\begin{equation}
Z' = f_{\bar{\theta}}(s_{t+\tau_{\mathrm{stop}}}), \quad D = \tau_{\mathrm{stop}}, \quad R = \sum_{k=0}^{D-1}\gamma^k r_{t+k}, \quad m = (1 - e_{t+D})\,m_{t+D}, \quad e = e_{t+D}.
\label{eq:fivetuple}
\end{equation}
Separating $m$ from $e$ is essential: internal skill termination ($m=1, e=0$) hands control back to the planner at $Z'$ with bootstrap value $V(Z')$, whereas episode absorption ($e=1$) zeroes future value identically. Defining the unified \emph{effective continuation gate} $a_1 \triangleq (1 - c_{\mathrm{mode}} m_1)(1 - e_1)$, concatenating two adjacent $h$-capped tuples $(Z'_1, D_1, R_1, m_1, e_1)$ and $(Z'_2, D_2, R_2, m_2, e_2)$ under identical $(\omega, h, c_{\mathrm{mode}})$ yields exact algebraic closure for the $2h$-capped tuple:
\begin{equation}
\begin{aligned}
Z'_{1\oplus 2} &= Z'_1 + a_1(Z'_2 - Z'_1), &\qquad D_{1\oplus 2} &= D_1 + a_1 D_2, \qquad R_{1\oplus 2} = R_1 + a_1 \gamma^{D_1} R_2, \\
m_{1\oplus 2} &= m_1 + (1 - m_1)(1 - e_1) m_2, &\qquad e_{1\oplus 2} &= e_1 + a_1 e_2.
\end{aligned}
\label{eq:semigroup}
\end{equation}
In the underlying Markov state space $\mathcal{S}$ under memoryless options $\pi_\omega(a\mid s)$, Equation~\eqref{eq:semigroup} holds as an exact pathwise identity (proof in Appendix~\ref{app:proofs}). Once projected into a compressed latent representation $z = f_\theta(s)$, semigroup closure relies on approximate latent value sufficiency \citep{grimm2020value}, motivating our decision value-equivalence consistency loss $\mathcal{L}_{\mathrm{decision}}$ matching single-$2h$ and composed $h+h$ option-Bellman backups (Figure~\ref{fig:arch}, Stage~2b).

\subsection{Marginal Covariance and Symlog Plug-In Bias vs.\ Endpoint Conditioning}

Given terminal value probe $V_{\bar{\xi}}: \mathcal{Z} \to \mathbb{R}$, the true stopped-option Bellman backup is $\mathcal{B}^\star(z,\omega) = \mathbb{E}[R + \tilde{\Gamma} V_{\bar{\xi}}(Z') \mid z,\omega,h]$ with effective discount $\tilde{\Gamma} = \gamma^D(1-e)$. Many multi-step models parameterize scalar heads marginally as $g_\psi(z,\omega,h)$ without $Z'$, and decode point predictions via 101-bin categorical two-hot cross-entropy in $\mathrm{symlog}(x) = \mathrm{sign}(x)\ln(1+|x|)$ space \citep{hafner2025mastering}, where point decoding $\mathcal{D}_{\mathrm{sym}}(p) = \mathrm{symexp}(\sum_{i=1}^{101} p_i b_i)$ recovers the inverse expected symlog transform.\footnote{Throughout all eight families, scalar heads decode via $\mathrm{symexp}(\sum_i p_i b_i)$ rather than linear bin expectation $\sum_i p_i\,\mathrm{symexp}(b_i)$, ensuring identical scalar parameterization across models and yielding the closed-form geometric-mean duration identity in Corollary~\ref{cor:symlog}.}

\begin{proposition}[Bifurcation Covariance Bias of Ideal Marginal Readouts]
\label{prop:bifurcation}
Suppose a world model matches the exact conditional endpoint law $Z' \mid (z,\omega,h)$. An ideal endpoint-conditioned readout predicting $\hat{R}(z,Z',\omega,h) = \mathbb{E}[R \mid z,Z',\omega,h]$ and $\hat{\Gamma}(z,Z',\omega,h) = \mathbb{E}[\tilde{\Gamma} \mid z,Z',\omega,h]$ yields an unbiased option-Bellman backup ($\mathcal{B}_{\mathrm{joint}} = \mathcal{B}^\star$), whereas an ideal marginal readout predicting $\bar{R} = \mathbb{E}[R \mid z,\omega,h]$ and $\bar{\Gamma} = \mathbb{E}[\tilde{\Gamma} \mid z,\omega,h]$ incurs exact bias
\begin{equation}
\mathcal{B}_{\mathrm{marg}}(z,\omega) - \mathcal{B}^\star(z,\omega) = -\mathrm{Cov}\!\left(\tilde{\Gamma},\, V_{\bar{\xi}}(Z') \,\middle|\, z, \omega, h\right).
\label{eq:covbias}
\end{equation}
\end{proposition}

In practice, planners evaluate a \emph{plug-in} discount $\gamma^{\hat{D}}(1-\hat{e})$ from categorical \texttt{Symlog-Twohot} point estimates rather than $\bar{\Gamma}$ directly. Accounting for the strict concavity of $\ln(1+D)$ on $D > 0$ yields an exact three-term decomposition that closes our analytical and learned corridor measurements to four decimal places:

\begin{corollary}[Three-Term Decomposition of Marginal \texttt{Symlog-Twohot} Plug-In Backups]
\label{cor:symlog}
Neglecting bin-discretization error on $[-20,20]$ for non-absorbing transitions ($e \equiv 0, D > 0$), cross-entropy-optimal marginal \texttt{Symlog-Twohot} decoding yields the shifted geometric mean $\hat{D}_{\mathrm{marg}} = \exp(\mathbb{E}[\ln(1+D)]) - 1 \le \mathbb{E}[D]$. Evaluating the planner plug-in backup $\widehat{\mathcal{B}}_{\mathrm{marg,plug}} = \hat{R}_{\mathrm{marg}} + \gamma^{\hat{D}_{\mathrm{marg}}}\mathbb{E}[V_{\bar{\xi}}(Z')]$ against $\mathcal{B}^\star$ decomposes exactly into
\begin{equation}
\small
\widehat{\mathcal{B}}_{\mathrm{marg,plug}} - \mathcal{B}^\star = \underbrace{\bigl(\hat{R}_{\mathrm{marg}} - \mathbb{E}[R]\bigr)}_{\text{(I) return symlog bias}} - \underbrace{\mathrm{Cov}\!\bigl(\gamma^D, V_{\bar{\xi}}(Z')\bigr)}_{\text{(II) discount--value cov.}} + \underbrace{\Bigl(\gamma^{\exp(\mathbb{E}[\ln(1+D)])-1} - \mathbb{E}[\gamma^D]\Bigr)\mathbb{E}[V_{\bar{\xi}}(Z')]}_{\text{(III) concave symlog duration plug-in shift}}.
\label{eq:three_term}
\end{equation}
Whenever the realized endpoint $Z'$ uniquely identifies the outcome branch ($\mathrm{Var}(D \mid z,Z',\omega,h) = \mathrm{Var}(R \mid z,Z',\omega,h) = 0$), conditioning $g_\psi$ on $Z'$ drives Terms~(I), (II), and~(III) simultaneously to zero (proof in Appendix~\ref{app:proofs}).
\end{corollary}

Conceptually, passing an endpoint into $g_\psi$ serves two distinct roles that must not be conflated:
\begin{enumerate}
\item \textbf{Sampled-Endpoint Conditioning} ($g_\psi(z, Z', \omega, h, c_{\mathrm{mode}})$): in stochastic multi-particle models, conditioning on each realized endpoint sample $Z'$ eliminates cross-branch covariance and duration plug-in bias (Corollary~\ref{cor:symlog}).
\item \textbf{Predicted-Endpoint Readout Anchoring} ($g_\psi(z, \mathrm{sg}(\hat{z}'), \omega, h, c_{\mathrm{mode}})$): during empirical training, supervising $g_\psi$ on the model's own stop-gradient generated endpoint $\hat{z}'$ aligns the readout's training support with the manifold actually queried during rollout and $\Delta_t$ bootstrapping.
\end{enumerate}

\subsection{Unified Training Protocol and Controlled Model Taxonomy}

All model families share an identical 64-dimensional encoder ($8\times 8$ \texttt{SimNorm} simplex product in the native benchmark), hidden width $256$, detached state-value probe $V_\xi(\mathrm{sg}(z))$, 101-bin \texttt{Symlog-Twohot} scalar heads, constant-rate AdamW ($3\times 10^{-4}$, batch size $128$, gradient clip $1.0$, EMA decay $0.999$), and 100,000-update budget. Algorithm~\ref{alg:biaxial} summarizes the bi-axial training step and chance-node MPC loop, and Table~\ref{tab:families} lists the architectural taxonomy across all benchmarked families and stabilized variants.

\begin{figure}[!t]
\refstepcounter{algorithmctr}\label{alg:biaxial}
\small
\begin{tabularx}{\textwidth}{@{\hspace{4pt}}X@{\hspace{4pt}}}
\labtoprule
\textbf{Algorithm \thealgorithmctr: Bi-Axial Semigroup Training (\texttt{JSK-Flow-Anchored}) and Chance-Node Jumpy MPC}\\\labheadrule
\textbf{Training Update} (batch size $128 = 96$ empirical rows $+ 32$ self-distillation rows; horizons $h \sim \mathcal{U}(\mathcal{H})$, mode $c_{\mathrm{mode}} \sim \mathcal{U}\{0,1\}$):\\
1.\ \emph{Empirical flow \& anchored readout (75\% batch)}: Encode $z = f_\theta(s_t)$, clean EMA target $z' = f_{\bar{\theta}}(s_{t+D})$. Compute conditional flow-matching loss $\mathcal{L}_{\mathrm{flow}}(z, z', \omega, h, c_{\mathrm{mode}})$ and generate $K=2$ endpoint $\hat{z}' = \Phi_\phi(z, \epsilon, \omega, h, c_{\mathrm{mode}})$. Supervise readout $g_\psi$ via $\mathcal{L}_{\mathrm{read}} = (1-\alpha)\,\mathcal{L}_{\mathrm{2hot}}(g_\psi(z, z'), y) + \alpha\,\mathcal{L}_{\mathrm{2hot}}(g_\psi(z, \mathrm{sg}(\hat{z}')), y)$ with internal weights $(1.0, 0.5, 0.5, 0.5)$ on $(R, D, m, e)$ ($\alpha=0$ in unanchored \texttt{JSK-Flow}; $\alpha=0.5$ in anchored variants).\\
2.\ \emph{Bi-axial self-distillation (25\% batch)}: Split equally into (a) $\Delta_\tau$ two-step shortcut consistency along $\tau \in [0,1]$, and (b) $\Delta_t$ dyadic semigroup composition ($h + h \to 2h$) sampling independent noises $\epsilon^{(1)} \perp \epsilon^{(2)} \sim \mathcal{N}(0,I)$, querying EMA teacher $(\Phi_{\bar{\phi}}, g_{\bar{\psi}})$ at $\hat{z}'_1$ and $\hat{z}'_2$ under $\mathrm{stop\_grad}$, composing $(Z'_{1\oplus 2}, D_{1\oplus 2}, R_{1\oplus 2}, m_{1\oplus 2}, e_{1\oplus 2})$ via Eq.~\eqref{eq:semigroup}, and adding decision value-equivalence $\mathcal{L}_{\mathrm{decision}}$ after 5,000 warmup updates.\\
3.\ \emph{Joint optimization}: Minimize $\mathcal{L} = \mathcal{L}_{\mathrm{emp}} + 0.25(\mathcal{L}_\tau + \mathcal{L}_\Delta + \mathcal{L}_{\mathrm{decision}}) + \mathcal{L}_{\mathrm{probe}}(\mathrm{sg}(z_t), G_t)$; clip $\|\nabla\|_2 \le 1.0$; update $(\bar{\theta}, \bar{\phi}, \bar{\psi}, \bar{\xi}) \leftarrow 0.999\,\bar{\Theta} + 0.001\,\Theta$.\\\labrowrule
\textbf{Chance-Node Receding-Horizon MPC} (state $s_0$, candidates $C=64$, particles $N_p$, per-candidate budget $B$, jump cap $h_{\mathrm{plan}}=32$):\\
1.\ Encode $z_0 = f_\theta(s_0)$ once; sample $C$ candidate sequences $\omega_{0:J-1}^{(c)}$ ($J = \lfloor B / K_{\mathrm{jump}} \rfloor$) and replicate each across $N_p$ independent particles.\\
2.\ Roll out $j = 0,\dots,J-1$ with $c_{\mathrm{mode}}=1$, accumulating particle return $\hat{Q}_{c,p} \leftarrow \hat{Q}_{c,p} + \Gamma_{c,p} \hat{R}_j$ and $\Gamma_{c,p} \leftarrow \Gamma_{c,p} \gamma^{\hat{D}_j}(1 - \hat{e}_j)$, plus terminal bootstrap $\Gamma_{c,p} V_{\bar{\xi}}(\hat{z}'_J)$. Average particles first, $\bar{Q}_c = \frac{1}{N_p}\sum_{p=1}^{N_p} \hat{Q}_{c,p}$, then select $c^\star = \arg\max_c \bar{Q}_c$ and execute $\omega_0^{(c^\star)}$ (1 step or until termination).\\\labrowrule
\end{tabularx}
\end{figure}

\begin{labtable}[!t]{Taxonomy of controlled model families and stabilized semigroup variants. ``Emp.\ Readout Target'' specifies the endpoint passed to $g_\psi$ on empirical batches ($z'$: clean EMA latent; $\mathrm{sg}(\hat{z}')$: stop-gradient predicted endpoint; ``None'': endpoint-free marginal). Lookahead denotes nominal environment steps at $B=8, h_{\mathrm{plan}}=32$.}{tab:families}
{>{\hsize=1.36\hsize\linewidth=\hsize\raggedright\arraybackslash}X
 >{\hsize=1.08\hsize\linewidth=\hsize\centering\arraybackslash}X
 >{\hsize=0.92\hsize\linewidth=\hsize\centering\arraybackslash}X
 >{\hsize=1.22\hsize\linewidth=\hsize\centering\arraybackslash}X
 >{\hsize=0.68\hsize\linewidth=\hsize\centering\arraybackslash}X
 >{\hsize=0.74\hsize\linewidth=\hsize\centering\arraybackslash}X}
\labh{Model Family} & \labh{Dynamics \& Space} & \labh{Horizon Rule} & \labh{Emp.\ Readout Target} & \labh{Solver K} & \labh{Lookahead}\\\labheadrule
Expectation-Option & Det.\ MLP (\texttt{SimNorm}) & Emp.\ + $\Delta_t$ & Predicted $\hat{z}'$ & 1 & 256\\\labrowrule
n-Step-MC & Flow map (\texttt{SimNorm}) & Emp.\ $h \le 128$ & Clean EMA $z'$ & 2 & 128\\\labrowrule
Geometric-TD-HC & Occ.\ flow (\texttt{SimNorm}) & Geometric TD & Horizon-cond.\ $z'$ & 2 & 128\\\labrowrule
Denoise-Only & Flow map (\texttt{SimNorm}) & Short $h \le 8$ ($4\times$) & Clean EMA $z'$ & $2\times 4$ & 32\\\labrowrule
JSK-Flow & Dual flow (\texttt{SimNorm}) & Emp.\ + $\Delta_t$ & Clean EMA $z'$ ($\alpha\!=\!0$) & 2 & 128\\\labrowrule
One-Step & Prim.\ MLP + $\hat{\pi}$ & $h=1$ rollout & Predicted $\hat{z}_{t+1}$ & 1 & 8\\\labrowrule
Time-Shortcut & Gaussian (\texttt{SimNorm}) & Emp.\ + $\Delta_t$ & Clean EMA $z'$ & 1 & 256\\\labrowrule
Marginal-Heads & Dual flow (\texttt{SimNorm}) & Emp.\ + $\Delta_t$ & None (marginal) & 2 & 128\\\labrowrule
JSK-Flow-Anchored & Dual flow (\texttt{SimNorm}) & Emp.\ + $\Delta_t$ & $0.5\,z' + 0.5\,\mathrm{sg}(\hat{z}')$ & 2 & 128\\\labrowrule
JSK-Flow-Sphere-Anc. & Geodesic flow ($\mathbb{S}^{63}(8)$) & Emp.\ + $\Delta_t$ & $0.5\,z' + 0.5\,\mathrm{sg}(\hat{z}')$ & 2 & 128\\
\end{labtable}

\section{What Drives Closed-Loop Jumpy Planning?}
\label{sec:experiments}

We evaluate across four environments spanning sparse bottleneck navigation (\textbf{\texttt{FourRooms}}; \citealp{sutton1999between}), sequential multi-stage pad activation (\textbf{\texttt{PinPad}}; \citealp{pasukonis2023evaluating,hafner2025mastering}), irreversible combinatorial box-pushing (\textbf{\texttt{Sokoban}}), and continuous-control locomotion (\textbf{\texttt{AntMaze}}; \citealp{park2025ogbench}), each paired with disjoint $2{,}000 / 200 / 200$ \texttt{Train} / \texttt{Val} / \texttt{Test} episode splits and two frozen option libraries ($|\Omega|=8$; complete physical dimensions and skill definitions in Appendix~\ref{app:setup}): low-coverage random linear controllers ($\mathcal{O}_{\mathrm{rand}}$, mean span $17.4\text{--}20.4$ steps, 0\% positive-reward episodes on \texttt{PinPad}/\texttt{Sokoban}) and macro-skills ($\mathcal{O}_{\mathrm{skill}}$, mean span $32.6\text{--}63.4$ steps). Across 8 native families $\times$ 4 tasks $\times$ 2 regimes $\times$ 3 seeds, we train 192 full-budget models and evaluate closed-loop MPC on both validation-selected (\texttt{Val-best}) and 100,000-update terminal checkpoints (Table~\ref{tab:main_mpc} and Figure~\ref{fig:tasks}).

\begin{labtable}[!t]{Closed-loop Test MPC solve rates (\%) at the default budget ($B=8, C=64, N_p=4$; 2,048 Dynamics Transition PNFE per decision; 1,200 Test episodes per task cell pooling primitive and option execution across 3 seeds). Under $\mathcal{O}_{\mathrm{skill}}$, cells report \textbf{\texttt{Val-best} / \texttt{100k}} solve rates side-by-side (across-seed SD for all cells is plotted in Figure~\ref{fig:tasks}). $^\dagger$\texttt{Time-Shortcut} 100k reflects 2 completed seeds on \texttt{PinPad} (seed 1 overflowed \texttt{float32} at update 82,706, whereas its saved 20k \texttt{Val-best} checkpoint completes all 3/3 seeds). Bottom rows report our stabilized 100k anchored variants on $\mathcal{O}_{\mathrm{skill}}$.}{tab:main_mpc}
{>{\hsize=1.52\hsize\linewidth=\hsize\raggedright\arraybackslash}X
 >{\hsize=0.90\hsize\linewidth=\hsize\centering\arraybackslash}X
 >{\hsize=0.92\hsize\linewidth=\hsize\centering\arraybackslash}X
 >{\hsize=0.88\hsize\linewidth=\hsize\centering\arraybackslash}X
 >{\hsize=0.88\hsize\linewidth=\hsize\centering\arraybackslash}X
 >{\hsize=0.90\hsize\linewidth=\hsize\centering\arraybackslash}X}
\labh{Regime / Model Family} & \labh{FourRooms} & \labh{PinPad} & \labh{Sokoban} & \labh{AntMaze} & \labh{Task Mean}\\\labheadrule
$\mathcal{O}_{\mathrm{rand}}$ (best: Expectation-Option) & 39.8 / 40.5 & 0.0 / 0.0 & 0.0 / 0.0 & 1.2 / 2.2 & 10.3 / 10.7\\\labrowrule
$\mathcal{O}_{\mathrm{rand}}$ (range, other 7 families) & 25.8\textendash 38.1 & 0.0 / 0.0 & 0.0 / 0.0 & 0.4\textendash 2.7 & 6.7\textendash 10.2\\\labrowrule
$\mathcal{O}_{\mathrm{skill}}$: Expectation-Option & 98.6 / 98.9 & 89.8 / \textbf{89.8} & 9.2 / 6.8 & \textbf{19.8} / \textbf{22.1} & \textbf{54.3} / \textbf{54.4}\\\labrowrule
$\mathcal{O}_{\mathrm{skill}}$: n-Step-MC & 99.2 / \textbf{99.6} & \textbf{90.1} / 89.4 & 2.1 / 5.3 & 15.8 / 18.7 & 51.8 / 53.3\\\labrowrule
$\mathcal{O}_{\mathrm{skill}}$: JSK-Flow (unanchored) & 97.0 / 79.7 & 79.1 / 60.8 & \textbf{10.5} / 3.6 & 6.3 / 6.0 & 48.2 / 37.5\\\labrowrule
$\mathcal{O}_{\mathrm{skill}}$: Geometric-TD-HC & 97.4 / 97.1 & 71.8 / 77.3 & 10.3 / \textbf{10.4} & 5.9 / 6.5 & 46.3 / 47.8\\\labrowrule
$\mathcal{O}_{\mathrm{skill}}$: Denoise-Only & 99.2 / 99.2 & 61.1 / 71.0 & 6.4 / 5.4 & 15.9 / 8.4 & 45.7 / 46.0\\\labrowrule
$\mathcal{O}_{\mathrm{skill}}$: Time-Shortcut$^\dagger$ & 92.8 / 91.8 & 41.1 / 39.5 & 6.9 / 5.8 & 11.8 / 7.9 & 38.1 / 35.9\\\labrowrule
$\mathcal{O}_{\mathrm{skill}}$: Marginal-Heads & 97.7 / 93.3 & 38.1 / 25.1 & 10.3 / 8.0 & 6.1 / 7.2 & 38.0 / 33.4\\\labrowrule
$\mathcal{O}_{\mathrm{skill}}$: One-Step & 96.5 / 94.8 & 38.6 / 39.2 & 6.2 / 5.7 & 5.4 / 5.8 & 36.7 / 36.4\\\labrowrule
$\mathcal{O}_{\mathrm{skill}}$: JSK-Flow-Sphere-Anc. & 93.4 / 91.5 & 73.2 / 63.2 & 8.8 / 2.8 & 9.8 / 7.1 & 46.3 / 41.2\\\labrowrule
$\mathcal{O}_{\mathrm{skill}}$: JSK-Flow-Anchored & \textbf{99.7} / 96.7 & 52.7 / 44.9 & 10.2 / 7.1 & 12.6 / 10.6 & 43.8 / 39.8\\
\end{labtable}

\subsection{Subgoal Reachability Governs Temporal Shortcuts at Matched Span Duration}

Table~\ref{tab:main_mpc} shows that temporal shortcuts cannot manufacture credit assignment when the option set never visits task subgoals: under $\mathcal{O}_{\mathrm{rand}}$, all eight families record 0.00\% across every \texttt{PinPad} and \texttt{Sokoban} episode, whereas $\mathcal{O}_{\mathrm{skill}}$ lifts \texttt{Expectation-Option} and \texttt{n-Step-MC} to 89.75\%--90.08\% on \texttt{PinPad} and 18.67\%--22.08\% on \texttt{AntMaze}. Comparing $\mathcal{O}_{\mathrm{rand}}$ and $\mathcal{O}_{\mathrm{skill}}$ alone, however, also changes option duration ($20.4 \to 60.9$ steps on \texttt{PinPad}). To isolate subgoal reachability from span length without retraining any weights, we exploit the internal structure of $\mathcal{O}_{\mathrm{skill}}$: options $0\text{--}3$ are sustained directional controllers (mean \texttt{PinPad} duration $55.3\text{--}58.8$ steps, 0\% pad completion), while options $4\text{--}7$ are subgoal-seeking skills with virtually identical duration ($51.6\text{--}59.8$ steps). Restricting the MPC action set across five nested levels $\texttt{L0}\text{--}\texttt{L4}$ unlocking $\{0, 1, 2, 3, 4\}$ subgoal skills (72,000 Test episodes; Figure~\ref{fig:coverage_frontier}a) holds realized \texttt{PinPad} option duration constant ($56.7 \to 57.0$ steps) while solve rates climb monotonically from 0.00\% (\texttt{L0}) to 83.83\% (\texttt{JSK-Flow} \texttt{Val-best}), 90.00\% (\texttt{Expectation-Option}), and 90.33\% (\texttt{n-Step-MC}), yielding four-task stratified paired $\texttt{L4} - \texttt{L0}$ gains of $+41.21$ percentage points (95\% CI $[39.42, 42.96]$) for \texttt{Expectation-Option} and $+43.38$ percentage points ($[40.17, 46.29]$) for \texttt{JSK-Flow}.

Skill statistics also explain why \texttt{Sokoban} peaks at 10.42\%--10.50\% despite a 48.0\% positive-reward trajectory rate (Appendix~\ref{app:setup}): pushing a \emph{single} box onto a target triggers shaped positive reward on 84\%--96\% of rollouts for options $4\text{--}7$, whereas full multi-box puzzle completion by a single option occurs on only 3.0\% of validation rollouts because irreversible corner deadlocks block remaining boxes. Online jumpy MPC thus lifts full-puzzle completion by 1.8--3.5$\times$ over single-option execution, while remaining bounded by irreversible deadlocks within fixed heuristic options.

\begin{labfigure}[!t]{Task-by-task closed-loop Test solve rates under $\mathcal{O}_{\mathrm{skill}}$ ($B=8, N_p=4$; solid bars show \texttt{Val-best} checkpoints and hatched bars show 100k terminal checkpoints, mean $\pm$ SD across 3 seeds). Validation checkpoint selection eliminates late-training semigroup degradation in \texttt{JSK-Flow}, lifting its four-task mean by $+10.71$ percentage points ($37.52\% \to 48.23\%$, rising from fifth to third overall behind \texttt{Expectation-Option} at 54.33\% and \texttt{n-Step-MC} at 51.77\%).}{fig:tasks}
\labincludegraphics{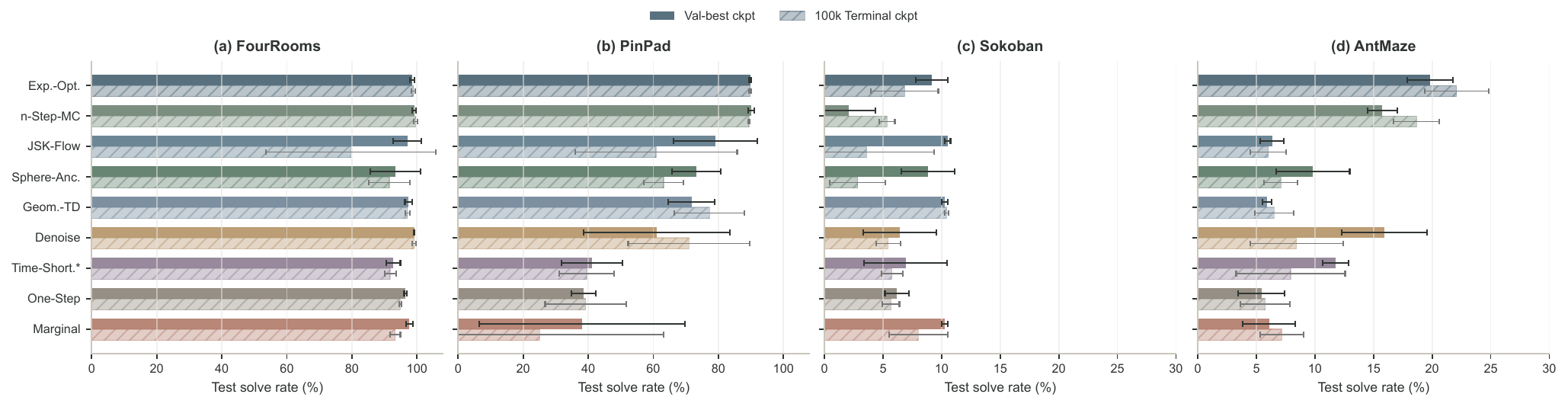}
\end{labfigure}

\subsection{Three-Tier Verification of Endpoint-Conditioned vs.\ Marginal Readouts}

We test Proposition~\ref{prop:bifurcation} and Corollary~\ref{cor:symlog} across three complementary settings that separate exact plug-in bias from end-to-end representation coupling (Table~\ref{tab:openloop_pred} and Figure~\ref{fig:bifurcation}):
\begin{enumerate}
\item \textbf{Exact Corridor Closure.} On a two-branch option ($p=0.5$: fast subgoal $D_{\mathrm{fast}}=4, V=1, R=0$; $p=0.5$: timeout $D_{\mathrm{slow}}=64, V=0, R=0$; $\gamma=0.99$), the true backup is $\mathcal{B}^\star = 0.5\times 0.99^4 = \mathbf{0.48030}$ and the ideal covariance-shifted marginal backup is $\mathbb{E}[\gamma^D]\mathbb{E}[V] = \mathbf{0.37155}$ (Term~II $= -0.10875$). Under \texttt{Symlog-Twohot}, marginal duration converges to $\hat{D}_{\mathrm{marg}} = \sqrt{5\times 65} - 1 = \mathbf{17.0278} \ll 34 = \mathbb{E}[D]$, shifting the plug-in backup by Term~III $= +0.04980$ to $0.99^{17.0278}\times 0.5 = \mathbf{0.42135}$—matching our trained marginal network ($\mathbf{0.42130}$, error $0.05900$) to $5\times 10^{-5}$, whereas the trained joint readout recovers $\mathbf{0.47957}$ ($81.4\times$ lower bias; Figure~\ref{fig:bifurcation}a).
\item \textbf{End-to-End Benchmark Backup Error and Frozen-Backbone Isolation.} Across the native benchmark, \texttt{Marginal-Heads} achieves near-best marginal Return MAE and Duration MAE yet inflates Option-Value Error by $22.6\times$ on $\mathcal{O}_{\mathrm{rand}}$ ($2.212$ vs.\ $0.098$) and $3.8\text{--}6.0\times$ on $\mathcal{O}_{\mathrm{skill}}$ ($3.659\text{--}5.529$ vs.\ $0.970\text{--}0.921$; Table~\ref{tab:openloop_pred}), collapsing \texttt{PinPad} MPC to 25.08\% (100k) and 38.08\% (\texttt{Val-best}). Because end-to-end training also shifts encoder latents ($\mathrm{MSE} = 0.0692$ vs.\ $0.0151$), we freeze converged \texttt{JSK-Flow-Anchored} backbones ($\Delta\mathrm{StateMSE} \equiv 0$, 18 readout-only runs) and train matched joint versus marginal heads with zero gradient into $(\theta,\phi)$ and no $\Delta_t$ feedback. On near-deterministic $\mathcal{O}_{\mathrm{skill}}$ trajectories where $(z,\omega,h)$ largely determines $Z'$, frozen joint and marginal heads perform comparably on \texttt{FourRooms}/\texttt{PinPad} ($|\Delta| \le 3.5$ percentage points) while joint conditioning reduces \texttt{AntMaze} option-value error ($0.545$ vs.\ $0.558$; $-8.1\%$ on high-variance doorway bifurcations, $0.205$ vs.\ $0.223$; Figure~\ref{fig:bifurcation}b). Thus, in end-to-end training on deterministic physics, omitting $Z'$ degrades performance primarily by destabilizing bootstrapped value targets and latent representations, whereas under genuinely bimodal transitions its irreducible covariance bias dominates directly.
\item \textbf{Bimodal Stochastic Control.} On environments with explicit risky-shortcut bifurcations (\texttt{StochFourRooms} and \texttt{StochPinPad}, Section~\ref{sec:regimes}), \texttt{Marginal-Heads} trails anchored spherical joint flows by $-29.33$ percentage points under option MPC ($95\%$ CI $[-48.25, -8.92]$; Figure~\ref{fig:bifurcation}c) and diverges by 60k updates ($2.65\times 10^4$ Return MAE).
\end{enumerate}

\begin{labtable}[!t]{Open-loop stopped-option prediction ($H=128$, four-task Test mean; values report \texttt{Val-best} $\to$ \texttt{100k} checkpoints, with Latent MSE reported at 100k). Normalized Option-Value Error is $\frac{1}{\sigma_{\mathrm{ret}}}\mathbb{E}[|\widehat{\mathcal{B}}_h^\omega V_{\bar{\xi}}(z) - (R + \gamma^D(1-e)V_{\bar{\xi}}(z'))|]$, using fixed $\mathcal{O}_{\mathrm{rand}}$ return scale $\sigma_{\mathrm{ret}}$.}{tab:openloop_pred}
{>{\hsize=1.58\hsize\linewidth=\hsize\raggedright\arraybackslash}X
 >{\hsize=0.92\hsize\linewidth=\hsize\centering\arraybackslash}X
 >{\hsize=0.94\hsize\linewidth=\hsize\centering\arraybackslash}X
 >{\hsize=0.84\hsize\linewidth=\hsize\centering\arraybackslash}X
 >{\hsize=0.72\hsize\linewidth=\hsize\centering\arraybackslash}X}
\labh{Regime / Family} & \labh{Return MAE} & \labh{Value Error} & \labh{Duration MAE} & \labh{Latent MSE}\\\labheadrule
$\mathcal{O}_{\mathrm{rand}}$: Expectation-Option & 0.089 $\to$ 0.089 & 0.103 $\to$ 0.098 & 12.16 $\to$ 12.22 & 0.0093\\\labrowrule
$\mathcal{O}_{\mathrm{rand}}$: n-Step-MC & \textbf{0.087} $\to$ \textbf{0.087} & \textbf{0.102} $\to$ \textbf{0.094} & 12.37 $\to$ 12.48 & \textbf{0.0089}\\\labrowrule
$\mathcal{O}_{\mathrm{rand}}$: Marginal-Heads & 0.093 $\to$ 0.095 & 2.938 $\to$ 2.212 & \textbf{12.14} $\to$ \textbf{12.18} & 0.0744\\\labrowrule
$\mathcal{O}_{\mathrm{rand}}$: JSK-Flow (unanchored) & 0.270 $\to$ $8.09\mathrm{e}7$ & 2.070 $\to$ $8.09\mathrm{e}7$ & 44.57 $\to$ 63.70 & 0.1103\\\labrowrule
$\mathcal{O}_{\mathrm{skill}}$: Expectation-Option & \textbf{0.711} $\to$ \textbf{0.717} & \textbf{0.970} $\to$ \textbf{0.921} & 31.22 $\to$ 31.89 & \textbf{0.0151}\\\labrowrule
$\mathcal{O}_{\mathrm{skill}}$: n-Step-MC & 0.714 $\to$ 0.727 & 1.257 $\to$ 1.062 & 32.04 $\to$ 32.92 & 0.0158\\\labrowrule
$\mathcal{O}_{\mathrm{skill}}$: Geometric-TD-HC & 0.763 $\to$ 0.792 & 1.389 $\to$ 1.445 & 55.58 $\to$ 72.72 & 0.0250\\\labrowrule
$\mathcal{O}_{\mathrm{skill}}$: Denoise-Only & 0.755 $\to$ 0.846 & 1.407 $\to$ 4.403 & 31.79 $\to$ 32.64 & 0.0356\\\labrowrule
$\mathcal{O}_{\mathrm{skill}}$: Time-Shortcut & \textbf{0.738} $\to$ 119.3 & 1.659 $\to$ 120.0 & \textbf{31.17} $\to$ 32.17 & 0.0272\\\labrowrule
$\mathcal{O}_{\mathrm{skill}}$: Marginal-Heads & 0.758 $\to$ 0.791 & 3.659 $\to$ 5.529 & 31.22 $\to$ \textbf{31.40} & 0.0692\\\labrowrule
$\mathcal{O}_{\mathrm{skill}}$: One-Step & 1.033 $\to$ 1.771 & 1.674 $\to$ 4.921 & 34.09 $\to$ 36.09 & 0.0799\\\labrowrule
$\mathcal{O}_{\mathrm{skill}}$: JSK-Flow (unanchored) & 1.075 $\to$ $1.51\mathrm{e}6$ & 1.999 $\to$ $1.51\mathrm{e}6$ & 53.63 $\to$ 70.39 & 0.1263\\
\end{labtable}

\begin{labfigure}[!t]{Verification of endpoint-conditioned joint kernels versus marginal readouts. \textbf{(a)} Exact three-term decomposition on the analytical corridor (Corollary~\ref{cor:symlog}), closing $\mathcal{B}^\star = 0.4803$, covariance bound $0.3716$, exact \texttt{Symlog} geometric-mean plug-in $0.4214$, learned marginal $0.4213$, and learned joint $0.4796$ ($81.4\times$ lower bias). \textbf{(b)} End-to-end benchmark Option-Value Error ($3.8\text{--}22.6\times$ inflation) alongside frozen-backbone readout isolation ($\Delta\mathrm{MSE} \equiv 0$). \textbf{(c)} Closed-loop Test solve rates on \texttt{PinPad} and bimodal stochastic control (\texttt{StochFourRooms}, \texttt{StochPinPad}), where anchored joint flows beat \texttt{Marginal-Heads} by $+29.3$ percentage points.}{fig:bifurcation}
\labincludegraphics{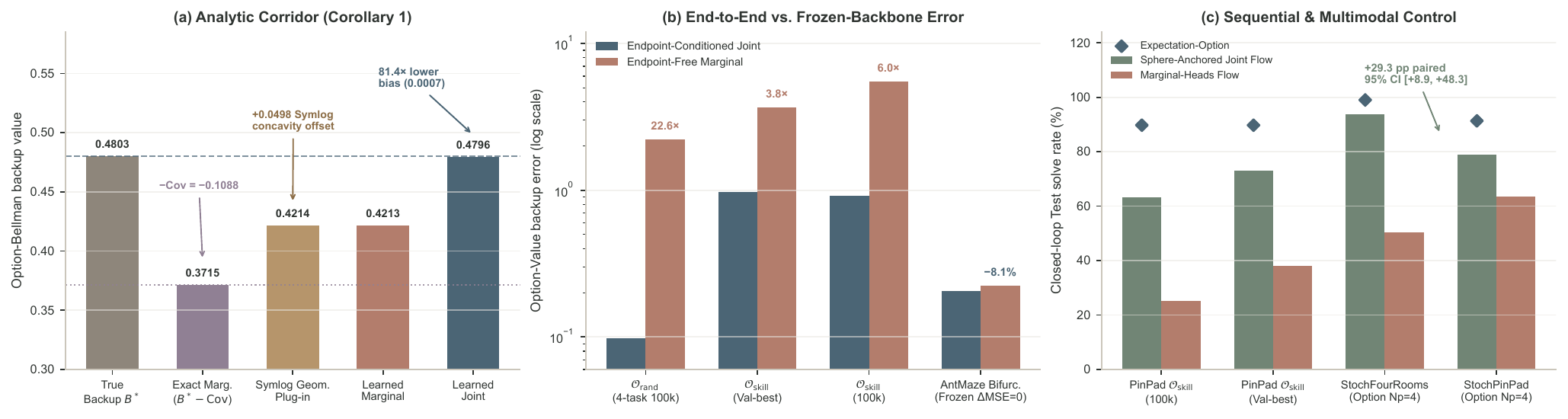}
\end{labfigure}

\subsection{Compute-Frontier Dominance, Validation Selection, and Matched Lookahead}

Across all six compute budgets ($B \in \{4,8,16\} \times N_p \in \{1,4\}$, 57,600 Test episodes per complete family; Figure~\ref{fig:coverage_frontier}b and Appendix~\ref{app:ledgers}), \texttt{Expectation-Option} ranks first overall ($18{,}777 / 57{,}600 = \mathbf{32.60\%}$ combined; $\mathbf{54.49\%}$ six-budget mean on $\mathcal{O}_{\mathrm{skill}}$), ahead of \texttt{n-Step-MC} ($18{,}115 / 57{,}600 = \mathbf{31.45\%}$ combined; $52.96\%$ on $\mathcal{O}_{\mathrm{skill}}$) and \texttt{Geometric-TD-HC} ($45.72\%$ on $\mathcal{O}_{\mathrm{skill}}$). At the minimum budget of \textbf{256 Dynamics Transition PNFE per decision} ($B=4, N_p=1$, 16.95 ms/decision), \texttt{Expectation-Option} reaches \textbf{54.75\%} on $\mathcal{O}_{\mathrm{skill}}$—exceeding both the strongest empirical flow baseline \texttt{n-Step-MC} (\textbf{53.29\%}, $+1.46$ percentage points) and the strongest bootstrapped flow baseline \texttt{Geometric-TD-HC} (\textbf{48.12\%}, $+6.63$ percentage points) at 4,096 PNFE ($16\times$ dynamics compression).

Evaluating all 192 validation-selected (\texttt{Val-best}) checkpoints across 76,800 Test episodes (Table~\ref{tab:main_mpc} and Figure~\ref{fig:tasks}) shows that early checkpoint selection leaves \texttt{Expectation-Option} firmly in first place (54.33\%) and \texttt{n-Step-MC} second (51.77\%), while rescuing unanchored \texttt{JSK-Flow} from late-stage readout saturation ($37.52\% \to \mathbf{48.23\%}$, $+10.71$ percentage points, rising to third overall). Furthermore, restricting single-step models to $J=4$ jumps (nominal cap 128, matching two-step flows at half their 1,024 vs.\ 2,048 dynamics PNFE across 172,800 episode-views; Figure~\ref{fig:coverage_frontier}c) yields \textbf{55.21\%} (\texttt{Val-best}) and \textbf{54.79\%} (100k) for \texttt{Expectation-Option}—outperforming its own $J=8$ setting (54.33\%) and every flow family at both $J=4$ and $J=8$ (4,096 PNFE). Thus its advantage stems from drift-free single-step endpoint prediction rather than extra nominal horizon.

Finally, committing to the selected option until termination rather than replanning every primitive step reduces episode-level PNFE by \textbf{25.5--27.2$\times$ pooled across tasks} ($\sum\mathrm{PNFE}_{\mathrm{prim}} / \sum\mathrm{PNFE}_{\mathrm{opt}}$: e.g., $1.015\times 10^6 \to 3.734\times 10^4$ on \texttt{Expectation-Option}; per-task pooled ratios: \texttt{FourRooms} $17.31\times$, \texttt{Sokoban} $25.16\times$, \texttt{AntMaze} $26.52\times$, \texttt{PinPad} $28.06\times$) and by \textbf{18.11--25.08$\times$ across 144 cell-level medians}, while \emph{increasing} $\mathcal{O}_{\mathrm{skill}}$ solve rates across all 8/8 families ($+1.54$ to $+18.32$ percentage points; Appendix~\ref{app:ledgers}).

\section{Resolving the Semigroup Manifold Trap}
\label{sec:trap}

\subsection{Exact Two-Hop Amplification Law on Unanchored Generated Supports}

Why does unanchored \texttt{JSK-Flow} saturate to $4.85\times 10^8$ Return MAE by 100k updates even though \texttt{SimNorm} strictly bounds every latent state to $\|z\|_2 \le \sqrt{8} \approx 2.83$—whereas \texttt{n-Step-MC} (same clean-endpoint readout, no $\Delta_t$ bootstrap) and \texttt{Expectation-Option} (same $\Delta_t$ bootstrap, readout anchored on $\hat{z}'$) remain completely stable? Figure~\ref{fig:mech_overview} illustrates the exact mechanism, formalized below.

\begin{labfigure}[!t]{Mechanistic anatomy of the unanchored semigroup manifold ratchet and dual stabilization. \textbf{(1)} On convex simplices $\Delta_7^8$, clean EMA encoder latents occupy low-entropy vertex neighborhoods ($H=0.398$), whereas two-step Euler flow endpoints land in the higher-entropy interior ($H=0.700$), remaining inside the domain ($2.4\times 10^{-7}$ violation) yet outside empirical readout support $\mathcal{M}_{\mathrm{clean}}$. \textbf{(2)} Because two-hop $\Delta_t$ composition queries $g_{\bar{\psi}}$ twice on $\mathcal{M}_{\mathrm{gen}}$, return bias amplifies by $(1 + a_1 \gamma^{D_1}) > 1$ per dyadic doubling ($\times 1.53$ even when $\gamma^{D_1} < 1$) until hitting $\mathrm{symexp}(20) \approx 4.85\times 10^8$; zero-retraining single-head swaps isolate this drift to the return head ($0.129$ MAE). \textbf{(3)} Stop-gradient readout anchoring on $\mathcal{M}_{\mathrm{gen}}$ and constant-norm Riemannian geodesic transport on $\mathbb{S}^{63}(8)$ eliminate all 100k divergences (0/27 anchored runs) and lift spherical MPC by $+20.7$ percentage points.}{fig:mech_overview}
\labincludegraphics{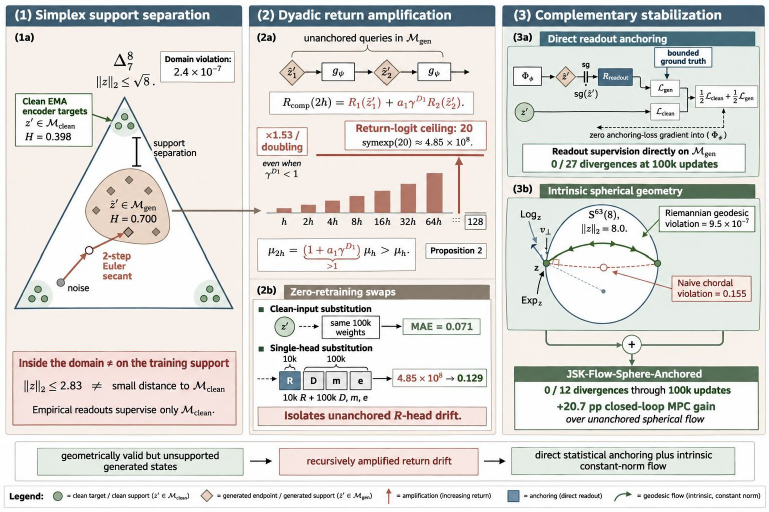}
\end{labfigure}

\begin{proposition}[Dyadic Return Amplification Under Support Separation and Dual Stabilization]
\label{prop:ratchet}
Let $\mathcal{M}_{\mathrm{clean}} \subset \mathcal{Z}$ denote the support of clean EMA latents $z' = f_{\bar{\theta}}(s')$ and $\mathcal{M}_{\mathrm{gen}} \subset \mathcal{Z}$ denote the support of finite-step flow endpoints $\hat{z}' = \Phi_\phi(z,\epsilon,\omega,h)$. On a bounded convex simplex product $\Delta_{m-1}^G$, Euclidean secant steps between vertex-concentrated codes satisfy $H(\hat{z}') > H(z')$, separating $\mathcal{M}_{\mathrm{gen}}$ into the interior away from $\mathcal{M}_{\mathrm{clean}}$.
\begin{enumerate}
\item \textbf{Super-unit dyadic amplification ($\gamma^{\hat{D}_1} < 1$ suffices):} Suppose empirical batches supervise $g_\psi$ solely on $\mathcal{M}_{\mathrm{clean}}$ ($\alpha=0$), while two-hop $\Delta_t$ composition queries EMA readouts at both generated hops $\hat{z}'_1, \hat{z}'_2 \in \mathcal{M}_{\mathrm{gen}}$:
\begin{equation}
\hat{R}_{\mathrm{comp}}(2h) = \hat{R}_{\bar{\psi}}(z, \hat{z}'_1, \omega, h) + \hat{a}_1 \gamma^{\hat{D}_1} \hat{R}_{\bar{\psi}}(\hat{z}'_1, \hat{z}'_2, \omega, h).
\end{equation}
If $\mathcal{M}_{\mathrm{gen}}$ exhibits positive readout drift $\mu_h > 0$ at scale $h$ and mean effective continuation-discount $\bar{\kappa}_h \triangleq \mathbb{E}[\hat{a}_1 \gamma^{\hat{D}_1}] > 0$, then even for $\gamma^{\hat{D}_1} < 1$, the composed $2h$ target drift satisfies
\begin{equation}
\mu_{2h} = (1 + \bar{\kappa}_h)\,\mu_h > \mu_h.
\label{eq:amplification}
\end{equation}
Without empirical restoring gradients on $\mathcal{M}_{\mathrm{gen}}$, compounding across $\log_2 H$ dyadic scales amplifies drift by $\prod_{\ell=0}^{\log_2 H - 1}(1 + \bar{\kappa}_{2^\ell}) > 1$ until $\hat{R}$ saturates at $R_{\max} = \mathrm{symexp}(20) \approx 4.85\times 10^8$ (producing an unclipped teacher target $(1 + \bar{\kappa}_h) R_{\max}$; at $\hat{D}_1 = 63.44, \hat{a}_1 \approx 1$, $1 + 0.99^{63.44} = 1.5285 \implies 7.416\times 10^8$).
\item \textbf{Complementary stabilization paths:} Breaking this feedback loop is achieved either statistically—by anchoring empirical readout supervision directly on $\mathrm{sg}(\hat{z}') \in \mathcal{M}_{\mathrm{gen}}$ ($\alpha > 0$)—or geometrically—by constraining $\mathcal{Z}$ to a constant-norm sphere $\mathbb{S}^{d-1}(\rho)$ with empty radial interior under intrinsic geodesic flow maps (proof in Appendix~\ref{app:proofs}).
\end{enumerate}
\end{proposition}

\subsection{Causal Isolation via Single-Head Swaps and Single-Factor Readout Anchoring}

Figure~\ref{fig:ratchet}a--c and Appendix~\ref{app:ledgers} verify every link of Proposition~\ref{prop:ratchet} without confounding solver steps or stochasticity:
\begin{itemize}
\item \textbf{Zero-Retraining Clean-Endpoint and Single-Head Swaps.} On \texttt{AntMaze} $\mathcal{O}_{\mathrm{rand}}$ seed 1, where 100k unanchored \texttt{JSK-Flow} reads $4.852\times 10^8$ on $\hat{z}'_{K=2}$ (with $\hat{m}_{K=2}$ collapsing from $0.945 \to 1.6\times 10^{-4}$ across updates 10k--100k while clean vs.\ generated SimNorm entropy separates at $0.398$ vs.\ $0.700$; Figure~\ref{fig:ratchet}b), feeding clean EMA targets $z'$ into the \emph{exact same 100k weights} immediately restores $0.0707$ Return MAE. Crucially, grafting \emph{only the 10k return head $R$} onto the 100k generator and 100k $(D,m,e)$ heads (\texttt{Swap-R-Only}, zero retraining) drops Return MAE from $4.852\times 10^8$ to $\mathbf{0.1287}$ (and on \texttt{AntMaze} $\mathcal{O}_{\mathrm{skill}}$ seed 2 from $57.60$ to $\mathbf{0.577}$, doubling option MPC solve rate from $9.5\%$ to $21.5\%$), whereas grafting 10k $(D,m,e)$ heads onto the 100k $R$ head fixes Duration MAE ($108.3 \to 15.3$) while leaving Return MAE at $4.852\times 10^8$ (Figure~\ref{fig:ratchet}c). Thus $R$-head drift on $\mathcal{M}_{\mathrm{gen}}$ is the primary driver, amplified by $(1 + \hat{a}_1\gamma^{\hat{D}_1})$.
\item \textbf{Single-Factor Predicted-Endpoint Readout Anchoring.} Holding $K=2$ Euler flow, \texttt{SimNorm}, and $\Delta_t$ bootstrapping fixed while setting $\alpha=0.5$ in Algorithm~\ref{alg:biaxial} (\texttt{JSK-Flow-Anchored}, 15 full 100k runs) achieves \textbf{0/15 divergences} at 100k updates—cutting \texttt{AntMaze} $\mathcal{O}_{\mathrm{rand}}$ terminal Return MAE from $1.62\times 10^8$ to $\mathbf{0.108}$ (with $\hat{m}_{128}$ locked at $0.998$; Figure~\ref{fig:ratchet}a,b), reducing $\mathcal{O}_{\mathrm{skill}}$ terminal Return MAE across all 4/4 tasks, and lifting \texttt{AntMaze} $\mathcal{O}_{\mathrm{skill}}$ 100k MPC from $6.00\%$ to $10.58\%$ ($12.58\%$ at \texttt{Val-best}; Table~\ref{tab:main_mpc}). A diagnostic $\alpha=1.0$ (\texttt{genread-only}) run likewise achieves 0 divergences ($0.104$ MAE, $18.0\%$ \texttt{AntMaze} option solve).
\end{itemize}

\begin{labfigure}[!t]{Quantitative causal dissection of semigroup divergence and dual stabilization. \textbf{(a)} Single-factor readout anchoring ($\alpha=0.5$, $K=2$ \texttt{SimNorm} flow unchanged) eliminates the nine-order-of-magnitude return ratchet through 100k updates, matching clean-endpoint evaluation. \textbf{(b)} Stacked evolution of normalized \texttt{SimNorm} entropy (top) and predicted stop probability $\hat{m}_{128}$ (bottom), showing persistent clean-vs.-generated entropy separation alongside anchored stop-gate stability ($\hat{m}_{128} \approx 0.998$). \textbf{(c)} Zero-retraining single-head checkpoint swaps isolating unanchored $R$-head drift ($0.129$ MAE). \textbf{(d)} Five-geometry 50k screen and 100k spherical verification (median bars, worst-case black diamonds, and 100k unanchored pure-sphere drift circle; \texttt{Unconstrained R64} domain violation is \texttt{N/A}).}{fig:ratchet}
\labincludegraphics{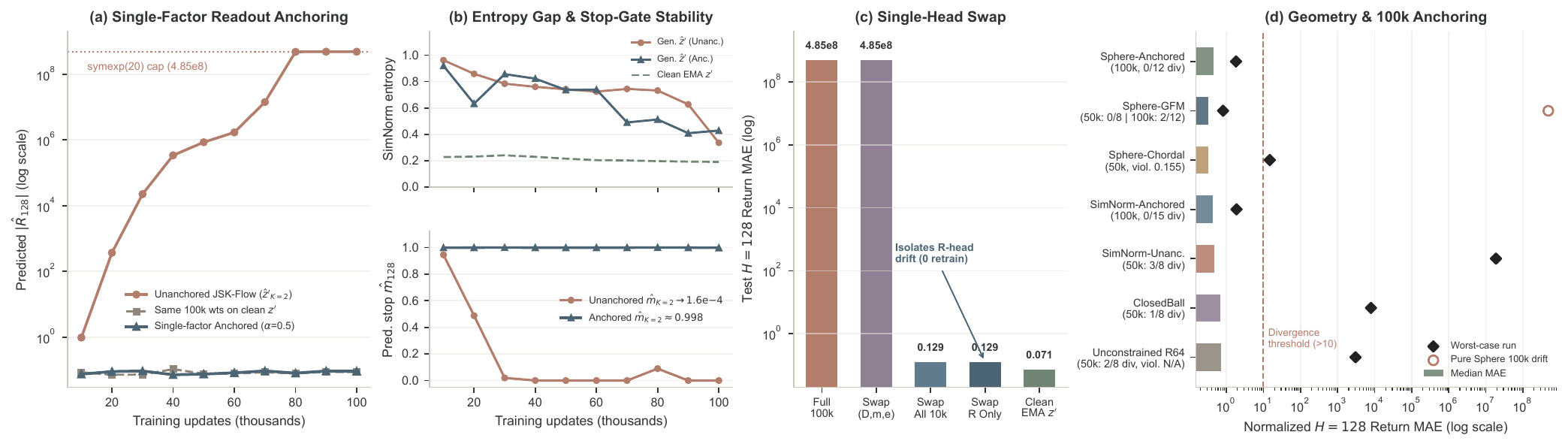}
\end{labfigure}

\subsection{Constant-Norm Spherical Flows at 50k and Synergy with Anchoring at 100k}

To test the geometric stabilization path, we first compare five latent geometries across 40 controlled 50,000-update runs ($\{\texttt{FourRooms}, \texttt{AntMaze}\} \times \{\mathcal{O}_{\mathrm{rand}}, \mathcal{O}_{\mathrm{skill}}\} \times 2$ seeds, part of 52 new 50k sensitivity/geometry training runs detailed in Appendix~\ref{app:setup}), then scale spherical geodesic flows to 100,000 updates with full closed-loop MPC across all four tasks ($\mathcal{O}_{\mathrm{skill}} \times 3$ seeds, 24 runs; Table~\ref{tab:geometry} and Figure~\ref{fig:ratchet}d).

\begin{labtable}[!t]{Latent geometry and readout-anchoring stabilization ledger ($H=128, K=2$). Top block: 50k screen across 8 conditions (values report median [worst-case max]; \textbf{Divergence} is defined as \texttt{NaN/Inf} training loss or Test normalized Return MAE $> 10.0$; \textbf{Violation} measures mean Euclidean midpoint distance to the target manifold, \texttt{N/A} on $\mathbb{R}^{64}$). Bottom block: full 100k four-task $\mathcal{O}_{\mathrm{skill}}$ closed-loop verification (12 runs per spherical arm; 15 runs for \texttt{SimNorm-Anchored}), reporting \texttt{Val-best} / \texttt{100k} MPC solve rate.}{tab:geometry}
{>{\hsize=1.64\hsize\linewidth=\hsize\raggedright\arraybackslash}X
 >{\hsize=0.92\hsize\linewidth=\hsize\centering\arraybackslash}X
 >{\hsize=0.90\hsize\linewidth=\hsize\centering\arraybackslash}X
 >{\hsize=0.74\hsize\linewidth=\hsize\centering\arraybackslash}X
 >{\hsize=0.80\hsize\linewidth=\hsize\centering\arraybackslash}X
 >{\hsize=1.00\hsize\linewidth=\hsize\centering\arraybackslash}X}
\labh{Geometry / Stabilization} & \labh{Return MAE} & \labh{Value Error} & \labh{Div.\ Rate} & \labh{Violation} & \labh{Step ms / MPC}\\\labheadrule
Sphere\_Riemannian\allowbreak\_GFM (50k) & \textbf{0.327} [0.817] & \textbf{0.602} [1.831] & \textbf{0 / 8} & $9.5\mathrm{e}{-7}$ & 48.4 ms\\\labrowrule
Sphere\_Naive\allowbreak\_Chordal (50k) & \textbf{0.327} [14.86] & 0.619 [16.50] & 1 / 8 & $1.55\mathrm{e}{-1}$ & 37.0 ms\\\labrowrule
SimNorm\_Convex (50k, $\alpha\!=\!0$) & 0.477 [$1.88\mathrm{e}7$] & 1.238 [$1.88\mathrm{e}7$] & 3 / 8 & $2.4\mathrm{e}{-7}$ & 33.1 ms\\\labrowrule
ClosedBall\_Convex (50k, $\mathbb{B}^{64}_8$) & 0.684 [7,973] & 1.130 [7,991] & 1 / 8 & 0.0 & 35.9 ms\\\labrowrule
Unconstrained\_R64 (50k) & 0.727 [3,058] & 2.192 [3,058] & 2 / 8 & N/A & \textbf{31.5 ms}\\\labrowrule
SimNorm-Anchored (100k, $\alpha\!=\!0.5$) & 0.434 [1.881] & \textbf{0.504} [9.452] & \textbf{0 / 15} & $2.4\mathrm{e}{-7}$ & 43.8\% / 39.8\%\\\labrowrule
Sphere-GFM (100k, $\alpha\!=\!0$) & 1.055 [$4.85\mathrm{e}8$] & 6.152 [$4.85\mathrm{e}8$] & 0/12 vb, 2/12 tm & $9.5\mathrm{e}{-7}$ & 25.6\% / 35.3\%\\\labrowrule
Sphere-Anchored (100k, $\alpha\!=\!0.5$) & \textbf{0.462} [\textbf{1.829}] & 2.635 [\textbf{6.729}] & \textbf{0 / 12} & $9.5\mathrm{e}{-7}$ & \textbf{46.3\%} / \textbf{41.2\%}\\
\end{labtable}

At 50k updates, every domain with a radial interior (\texttt{SimNorm}, \texttt{ClosedBall}) or unbounded norm ($\mathbb{R}^{64}$) suffers catastrophic divergence, and ambient chordal sphere projection incurs a $14.86$ spike from off-manifold midpoints ($0.155$ violation), whereas intrinsic geodesic transport on $\mathbb{S}^{63}(8)$ (\texttt{Sphere\_Riemannian\_GFM}) achieves 0/8 divergences (worst-case MAE $0.817$). Scaling unanchored spherical geodesic flow to 100k across all four tasks reveals a subtle geometric insight: while eliminating radial interior drift yields 0/12 divergences at \texttt{Val-best}, 2/12 runs (\texttt{FourRooms} seeds 0 and 1) still experience late 80k--100k \emph{tangential angular support drift} on $\mathbb{S}^{63}(8)$ (coordinate-energy entropy $H(p)/\ln 64$ shifting by $+0.076$ between clean and generated latents). Combining constant-norm spherical transport with $\alpha=0.5$ anchoring (\textbf{\texttt{JSK-Flow-Sphere-Anchored}}) closes both radial and tangential gaps—achieving \textbf{0/12 divergences across 50k, \texttt{Val-best}, and 100k checkpoints} (0/27 divergences across all anchored 100k runs) and lifting four-task \texttt{Val-best} closed-loop MPC by \textbf{$+20.69$ percentage points} over unanchored spherical flow ($25.62\% \to 46.31\%$; \texttt{FourRooms} $49.08\% \to 93.42\%$, \texttt{PinPad} $37.58\% \to 73.17\%$, \texttt{AntMaze} $6.17\% \to 9.83\%$).

\section{Horizon Extrapolation and Multimodal Stochastic Regimes}
\label{sec:regimes}

Why retain generative semigroup flows at all if deterministic \texttt{Expectation-Option} dominates our native benchmark? Our native tasks satisfy two conditions favorable to deterministic empirical models: offline trajectories contain full $H=128$ spans, and environment transitions given $(s,\omega)$ are near-deterministic. We test what happens when each assumption is relaxed across 36 controlled runs (Table~\ref{tab:regimes} and Figure~\ref{fig:regimes}).

\begin{labfigure}[!t]{Regime boundaries separating deterministic expectation models, empirical flow maps, and anchored spherical semigroup flows. \textbf{(a)} Trained strictly on truncated spans $h \le 16$, \texttt{JSK-Flow-Sphere-Anchored} achieves the lowest $8\times$ extrapolated open-loop Return MAE at $H=128$ on \texttt{FourRooms} (0.193 vs.\ 0.233 for \texttt{Expectation-Option} and 0.331 for non-bootstrapped \texttt{n-Step-MC}). \textbf{(b)} In closed-loop option MPC at equalized 2,048 PNFE under $h \le 16$, chaining two in-distribution $h=16$ jumps (\texttt{n-Step-MC}) and \texttt{Expectation-Option} remain strong on sparse \texttt{PinPad} because multi-stage dyadic extrapolation compounds variance on rare rewards. \textbf{(c)} On bimodal risky-shortcut environments (\texttt{StochFourRooms}, \texttt{StochPinPad}), \texttt{Sphere-Anchored} unlocks $+10.2$ percentage points from multi-particle ($N_p=4$) primitive replanning and beats \texttt{Marginal-Heads} by $+29.3$ percentage points under option execution.}{fig:regimes}
\labincludegraphics{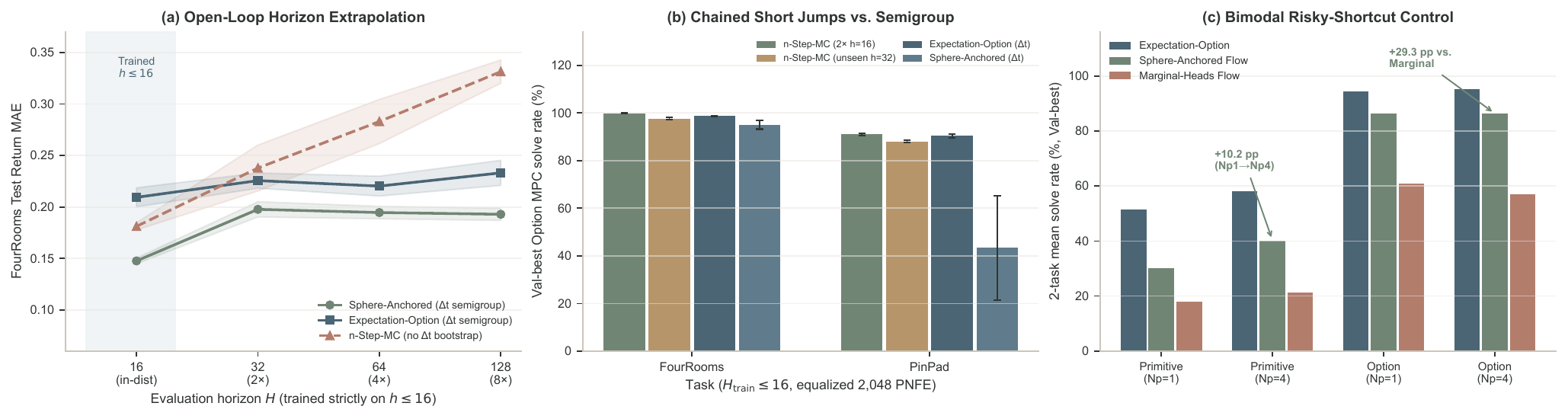}
\end{labfigure}

\textbf{Truncated-Horizon Extrapolation ($H_{\mathrm{train}} \le 16$).} Restricting offline supervision strictly to $h \in \{1,2,4,8,16\}$ across \texttt{FourRooms} and \texttt{PinPad} (18 runs to 100k updates) forces models to extrapolate $8\times$ beyond training spans at $H=128$. Without $\Delta_t$ bootstrapping, \texttt{n-Step-MC} degrades sharply on \texttt{FourRooms} ($H=128$ Return MAE $0.331$, semigroup self-consistency residual $0.201$), whereas \texttt{JSK-Flow-Sphere-Anchored} achieves the \textbf{lowest extrapolated open-loop error ($\mathbf{0.193}$, residual $\mathbf{0.027}$)}, ahead of \texttt{Expectation-Option} ($0.233$; Figure~\ref{fig:regimes}a). In closed-loop MPC at equalized 2,048 PNFE, however, a planner does not need a single monolithic jump: chaining two in-distribution $h=16$ flow jumps (\texttt{n-Step-MC} $2\times 16$: $99.83\%$ \texttt{FourRooms}, $91.00\%$ \texttt{PinPad}) and \texttt{Expectation-Option} ($98.67\%$ / $90.33\%$) outperform extrapolated \texttt{Sphere-Anchored} on sparse \texttt{PinPad} ($95.00\%$ / $43.33\%$, paired two-task difference $-26.25$ percentage points; Figure~\ref{fig:regimes}b), where three dyadic doublings ($16 \to 32 \to 64 \to 128$) compound variance across rare pad-completion sequences (100k \texttt{PinPad} MAE $10.62$).

\textbf{Bimodal Risky-Shortcut Bifurcations.} In \texttt{StochFourRooms} and \texttt{StochPinPad} (18 runs to 60k updates), risky doorway/pad shortcuts succeed rapidly ($p=0.5$, short $D$, high $V$) or detour into long recovery paths ($p=0.5$, long $D$, low $V$). Here \texttt{JSK-Flow-Sphere-Anchored} achieves the \textbf{lowest open-loop $H=128$ Return MAE on both tasks} ($0.299$ and $0.181$ at \texttt{Val-best}; $0.293$ and $0.184$ at 60k vs.\ $0.338$ and $0.206$ for \texttt{Expectation-Option}), gains \textbf{$+10.17$ percentage points} ($95\%$ CI $[+4.42, +15.83]$) when scaling particles from $N_p=1$ to $N_p=4$ under primitive replanning, and outperforms \texttt{Marginal-Heads} by \textbf{$+29.33$ percentage points} ($95\%$ CI $[+8.92, +48.25]$) under option MPC (Figure~\ref{fig:regimes}c). Even here, deterministic \texttt{Expectation-Option} paired with option-to-termination execution remains remarkably resilient ($95.17\%$ vs.\ $86.33\%$ two-task option mean) because committing to full macro-options smooths within-option branching noise between replanning points.

\begin{labtable}[!t]{Summary of regime-boundary experiments (\texttt{Val-best} Test metrics across 3 seeds per task). Left columns: $8\times$ open-loop extrapolation ($H=128$) and option MPC ($B=8, N_p=4$) when trained strictly on $h \le 16$. Right columns: bimodal risky-shortcut control (\texttt{StochFourRooms} and \texttt{StochPinPad} two-task means).}{tab:regimes}
{>{\hsize=1.54\hsize\linewidth=\hsize\raggedright\arraybackslash}X
 >{\hsize=0.88\hsize\linewidth=\hsize\centering\arraybackslash}X
 >{\hsize=0.90\hsize\linewidth=\hsize\centering\arraybackslash}X
 >{\hsize=0.88\hsize\linewidth=\hsize\centering\arraybackslash}X
 >{\hsize=0.90\hsize\linewidth=\hsize\centering\arraybackslash}X
 >{\hsize=0.90\hsize\linewidth=\hsize\centering\arraybackslash}X}
\labh{Model Variant} & \labh{FR $H\!=\!128$ MAE} & \labh{Short16 MPC} & \labh{Stoch MAE} & \labh{Prim $N_p$: 1$\to$4} & \labh{Stoch Opt $N_p\!=\!4$}\\\labheadrule
Expectation-Option & 0.233 & 94.50\% & 0.269 & 51.3\% $\to$ 58.2\% & \textbf{95.17\%}\\\labrowrule
JSK-Flow-Sphere-Anc. & \textbf{0.193} & 69.17\% & \textbf{0.240} & 30.1\% $\to$ 40.3\% & 86.33\%\\\labrowrule
n-Step-MC ($2\times 16$ / ood32) & 0.331 & \textbf{95.42\%} / 92.83\% & \textemdash & \textemdash & \textemdash\\\labrowrule
Marginal-Heads (bimodal) & \textemdash & \textemdash & 1.467 ($2.7\mathrm{e}4$ 60k) & 17.9\% $\to$ 21.3\% & 57.00\%\\
\end{labtable}

\section{Discussion, Design Playbook, and Broader Implications}
\label{sec:discussion}

\subsection{A Practitioner's Playbook: Which Jumpy World Model Should You Build?}

Because prior studies co-designed option libraries, generative transport, scalar parameterizations, and planning budgets simultaneously, practitioners lacked a clear decision rule for choosing among deterministic expectation jumps, direct empirical flow maps, and semigroup-bootstrapped flows. Table~\ref{tab:playbook} distills our 319-model study into an operational design prescription across four recurring problem regimes.

\begin{labtable}[!t]{Practitioner's design playbook for temporally abstract world models under fixed test-time compute, synthesized from 319 trained models and over 770,000 closed-loop evaluation episodes.}{tab:playbook}
{>{\hsize=0.92\hsize\linewidth=\hsize\raggedright\arraybackslash}X
 >{\hsize=0.92\hsize\linewidth=\hsize\raggedright\arraybackslash}X
 >{\hsize=1.06\hsize\linewidth=\hsize\raggedright\arraybackslash}X
 >{\hsize=1.10\hsize\linewidth=\hsize\raggedright\arraybackslash}X}
\labh{Problem Regime} & \labh{Recommended Model} & \labh{Key Ingredients} & \labh{Empirical Payoff}\\\labheadrule
Full-span logs ($H_{\mathrm{train}} \ge h_{\mathrm{plan}}$), near-deterministic & Expectation-Option ($K=1$, det.\ MLP) & Five-tuple $(Z',D,R,m,e)$; readout on $\hat{z}'$; option execution & 54.75\% at 256 PNFE (16.95 ms); $>400\times$ fewer dynamics calls/ep than step-level One-Step\\\labrowrule
Short-span logs ($H_{\mathrm{train}} \le 16$), open-loop long-horizon value & JSK-Flow-Sphere-Anc.\ ($K=2$, $\Delta_t$ semigroup) & $\mathbb{S}^{d-1}(\rho)$ geodesic flow; $\alpha=0.5$ $\mathrm{sg}(\hat{z}')$ anchor; $\mathcal{L}_{\mathrm{decision}}$ & Lowest $8\times$ extrapolated $H=128$ error (0.193 vs.\ 0.331 n-Step-MC); $7.4\times$ lower residual\\\labrowrule
Short-span logs ($H_{\mathrm{train}} \le 16$), sparse-reward closed-loop MPC & Chained n-Step-MC or Expectation-Option & Chain $2$\textendash $4$ in-distribution jumps ($h=16$) vs.\ $8\times$ jump & 95.42\% (n-Step-MC $2\times 16$) and 94.50\% (Exp.-Opt.), avoiding multi-stage variance\\\labrowrule
Multimodal / risky-shortcut stochastic dynamics & JSK-Flow-Sphere-Anc.\ + Chance-Node ($N_p \ge 4$) & Joint $(z, Z', \omega, h)$ heads; $\mathbb{S}^{d-1}(\rho)$; $\alpha=0.5$ anchor; particle mean & Lowest bimodal MAE (0.240); $+29.3$ pp over Marginal-Heads; $+10.2$ pp from $N_p=1 \to 4$\\
\end{labtable}

Two practical insights stand out. First, \textbf{simplicity wins whenever full option spans are observable offline and environment physics are near-deterministic}: a single-step ($K=1$) deterministic expectation jump with predicted-endpoint readout anchoring avoids flow-integration drift entirely, halves per-jump NFE cost, and—when paired with option-to-termination execution—combines $16\times$ per-decision dynamics compression with $25.5\text{--}27.2\times$ episode-level execution amortization ($>400\times$ reduction in dynamics evaluations per episode relative to primitive step-by-step replanning). Second, \textbf{when transitions genuinely bifurcate or long-horizon distributions must be extrapolated beyond truncated logs, every piece of our stabilized semigroup stack becomes load-bearing}: omitting $Z'$ from readouts triggers covariance and plug-in collapse ($-29.3$ percentage points), omitting readout anchoring or constant-norm spherical geometry reintroduces late-stage bootstrap divergence, and evaluating single particles sacrifices $+10.2$ percentage points of closed-loop risk awareness under step-level replanning.

\subsection{Three Broader Lessons for Self-Predictive World Models}

Beyond option-level planning, our causal diagnostics clarify three structural phenomena that apply broadly to self-supervised and generative world modeling:

\textbf{1.\ Norm Boundedness Does Not Imply Support Alignment.} Projecting latent representations onto bounded convex simplices (\texttt{SimNorm}) has become a standard recipe for preventing representation collapse without observation reconstruction \citep{lavoie2023simplicial,hansen2024tdmpc2}. Our fixed-weight counterfactual swaps and single-head grafts reveal why this guarantee breaks as soon as world models move from deterministic one-step rollouts to few-step generative transport: convex combinations of boundary-peaked codes remain strictly inside the norm bound ($2.4\times 10^{-7}$ domain violation) while migrating into a higher-entropy interior ($H: 0.398 \to 0.700$) where empirical readouts receive zero supervision. Even on a constant-norm hypersphere $\mathbb{S}^{63}(8)$ with empty radial interior, where intrinsic Riemannian geodesic flows eliminate all divergences through 50k updates and \texttt{Val-best}, 2 of 12 unanchored 100k runs still develop late tangential angular drift ($\Delta H = +0.076$). Statistical anchoring on stop-gradient generated latents ($\mathrm{sg}(\hat{z}')$) and intrinsic constant-norm geometry are therefore complementary rather than redundant—jointly achieving 0 divergences across all 27 anchored 100k runs.

\textbf{2.\ Categorical Scalar Codecs Interact Nonlinearly with Variable-Horizon Discounting.} Training value and reward heads via \texttt{Symlog-Twohot} cross-entropy is widely treated as a drop-in scale-invariant replacement for mean-squared regression \citep{farebrother2024stop,hafner2025mastering}. Corollary~\ref{cor:symlog} shows that under stochastic multi-step durations, point decoding $\mathrm{symexp}(\sum_i p_i b_i)$ recovers the shifted geometric mean $\exp(\mathbb{E}[\ln(1+D)]) - 1$ rather than $\mathbb{E}[D]$ or $-\ln\mathbb{E}[\gamma^D]/(-\ln\gamma)$, introducing a systematic concave duration shift alongside $-\mathrm{Cov}(\tilde{\Gamma}, V(Z'))$. Conditioning all scalar heads jointly on $(z, Z', \omega, h)$ resolves both biases simultaneously without abandoning scale-invariant categorical classification.

\textbf{3.\ Test-Time Compute Scaling Stalls Without Temporal Abstraction.} Increasing test-time search compute is often expected to monotonically improve control \citep{zhang2026knowing}. Yet in our six-budget sweep (Appendix~\ref{app:ledgers}), scaling primitive \texttt{One-Step} planning $16\times$ from 256 to 4,096 dynamics evaluations per decision yields \emph{zero} improvement on $\mathcal{O}_{\mathrm{skill}}$ ($37.35\% \to 36.54\%$), because expanding more trajectories of depth $8$ still cannot bridge $50\text{--}60$ step credit-assignment gaps on \texttt{PinPad} or \texttt{AntMaze}. By contrast, pairing subgoal-covering options with jumpy outcome prediction turns the smallest budget (256 evaluations, 16.95 ms) into 54.75\% solve rate. Test-time compute scaling in sequential decision-making is thus fundamentally gated by the temporal horizon per function evaluation.

\subsection{Scope Boundaries and Open Research Frontiers}
\label{sec:limitations}

Our study also sharpens three open frontiers for temporally abstract world models:
\begin{itemize}
\item \textbf{Closing the Loop with Online Option Discovery.} Freezing option libraries offline isolates world-model design cleanly, but caps performance on irreversible combinatorial puzzles such as \texttt{Sokoban} at 10.50\%, where heuristic skills cannot recover once a box is pushed into a corner deadlock. Co-evolving option policies and termination conditions online using value gradients from the stopped-option world model is a natural step toward realizing the full Options-and-Knowledge loop \citep{sutton1999between}.
\item \textbf{Uncertainty-Adaptive Jump Horizons.} Under truncated offline supervision ($H_{\mathrm{train}} \le 16$) on sparse sequential rewards (\texttt{PinPad}), chaining two in-distribution $h=16$ jumps markedly outperforms extrapolating three dyadic doublings to $H=128$ ($91.00\%$ vs.\ $43.33\%$). Equipping planners with epistemic uncertainty estimates over $\Delta_t$ composition would allow agents to leap coarsely ($h=64\text{--}128$) across open corridors while automatically stepping finely ($h=4\text{--}16$) near sparse bottlenecks.
\item \textbf{Scaling Spherical Stopped-Option Semigroups to Visual World Models.} All experiments here use structured state observations to permit 319 controlled training runs and over 770,000 closed-loop evaluation episodes (alongside minor numerical safeguards such as clamping pre-projection Gaussian log-variance $\log\sigma \in [-10, 2]$ to avoid the single \texttt{float32} overflow seen in \texttt{Time-Shortcut} at 82.7k updates). Extending five-tuple stopped-option semigroups and constant-norm spherical flows to high-dimensional visual and video world models—where skipping frame-by-frame pixel synthesis could yield order-of-magnitude wall-clock speedups—is an exciting direction for foundation world models.
\end{itemize}

\section{Conclusion}

Escaping the one-step trap in model-based reinforcement learning requires more than replacing one-step MLPs with modern generative flows; it requires rethinking how temporal abstraction, value backups, and latent geometry interact under a fixed test-time compute budget. By formulating multi-step skill dynamics as a five-tuple stopped-option semigroup $(Z', D, R, m, e)$ that distinguishes internal skill termination from episode absorption, we placed deterministic expectation models, empirical shortcut flow maps, occupancy flows, and bi-axial semigroup flows on a common mathematical and experimental footing across 319 trained models and more than 770,000 closed-loop planning episodes.

Four durable principles emerge from this synthesis. First, temporal shortcuts are amplifiers of behavioral subgoal reachability rather than substitutes for it—a law that holds even within a single skill library when option durations are strictly matched. Second, variable-duration jumps require endpoint-conditioned joint readouts: omitting the realized endpoint $Z'$ entangles fast subgoal arrivals with slow timeouts through both discount--value covariance and concave \texttt{Symlog} duration plug-in bias. Third, bounded convex latent spaces harbor an interior support gap under finite-step generative transport that turns two-hop semigroup bootstrapping into a $(1 + \hat{a}_1\gamma^{\hat{D}_1}) > 1$ return ratchet—diagnosed without retraining via single-head swaps and eliminated across all 27/27 anchored full-budget runs through stop-gradient predicted-endpoint readout anchoring and constant-norm Riemannian spherical geodesic flows. Fourth, deterministic single-step expectation jumps and multi-particle spherical semigroup flows occupy complementary halves of the problem landscape: expectation models paired with option-to-termination execution dominate full-span near-deterministic control with over $400\times$ fewer dynamics evaluations per episode than primitive step-by-step planning, whereas anchored spherical semigroup flows unlock long-horizon extrapolation from truncated logs and risk-aware planning under multimodal stochastic dynamics.

Taken together, these results replace ad-hoc co-design in temporally abstract world modeling with a simple, verifiable engineering blueprint. Moving world models from simulating clock ticks to composing stopped skills transforms how test-time compute translates into long-horizon intelligence—laying a rigorous foundation for agents that plan in leaps rather than steps.

\begin{ack}
\labstandardack
\end{ack}

\bibliography{refs}

\appendix

\section{Mathematical Proofs}
\label{app:proofs}

\subsection{Exact Pathwise Semigroup Closure of Stopped-Option Five-Tuples}

Fix any trajectory $(s_t, s_{t+1}, \dots)$ under Markov option $\omega$, horizon cap $h \ge 1$, and mode $c_{\mathrm{mode}} \in \{0,1\}$. Define stopping sets $\mathcal{S}_1 = \{k \in \{1,\dots,h\} : e_{t+k}=1 \lor (c_{\mathrm{mode}}=1 \land m_{t+k}=1)\}$ on $[t+1, t+h]$ and $\mathcal{S}_2 = \{j \in \{1,\dots,h\} : e_{t+h+j}=1 \lor (c_{\mathrm{mode}}=1 \land m_{t+h+j}=1)\}$ on $[t+h+1, t+2h]$, with non-absorbing skill-stop indicator $m_i = (1 - e_{t+\tau_i}) m_{t+\tau_i}^{\mathrm{raw}} \in \{0,1\}$ and effective continuation gate $a_1 = (1 - c_{\mathrm{mode}} m_1)(1 - e_1) \in \{0,1\}$.

\textbf{Case 1 ($\mathcal{S}_1 \neq \emptyset$):} Let $\tau_1 = \min \mathcal{S}_1 \le h$. Either $e_1 = 1$ (giving $a_1 = 0$ and $(1 - m_1)(1 - e_1) = 0$) or $e_1 = 0$ with $c_{\mathrm{mode}} = 1$ and $m_1 = 1$ (giving $a_1 = (1 - 1)(1 - 0) = 0$ and $(1 - m_1)(1 - e_1) = 0$). Since $\mathcal{S}_1 \subset \{1,\dots,2h\}$ is non-empty, the $2h$-capped stopping index is also $\tau_1$, and substituting $a_1 = 0$ and $(1-m_1)(1-e_1)=0$ into Eq.~\eqref{eq:semigroup} yields $(Z'_1, D_1, R_1, m_1, e_1)$, matching the $2h$-tuple identically.

\textbf{Case 2 ($\mathcal{S}_1 = \emptyset$):} Then $D_1 = h$, $e_1 = 0$, and either $c_{\mathrm{mode}} = 0$ (fixed-span mode, where $m_1 = m_{t+h}$ may be $0$ or $1$ while $c_{\mathrm{mode}} m_1 = 0 \implies a_1 = 1$; note for $c_{\mathrm{mode}}=0$ the composed internal flag $m_{1\oplus 2}$ is unused by continuation gating, and for $c_{\mathrm{mode}}=1$ we have $m_1 = 0$) or $c_{\mathrm{mode}} = 1$ with $m_1 = 0$, giving $a_1 = 1$ and $(1 - m_1)(1 - e_1) = 1$. The $2h$-stopping index is $h + D_2$, so $Z'_{2h} = Z'_2 = Z'_1 + a_1(Z'_2 - Z'_1)$, $D_{2h} = h + D_2 = D_1 + a_1 D_2$, $e_{2h} = e_2 = e_1 + a_1 e_2$, $m_{2h} = m_2 = m_1 + (1-m_1)(1-e_1)m_2$ (under $c_{\mathrm{mode}}=1$), and $R_{2h} = \sum_{k=0}^{h-1}\gamma^k r_{t+k} + \gamma^h \sum_{j=0}^{D_2-1}\gamma^j r_{t+h+j} = R_1 + a_1 \gamma^{D_1} R_2$. In Markov state space $\mathcal{S}$ with memoryless $\pi_\omega(a\mid s)$, the conditional law of tuple 2 given $(s_{t+h}, \omega, h, c_{\mathrm{mode}})$ depends only on $s_{t+h}$; in latent space $z = f_\theta(s)$, distributional semigroup closure holds up to the latent Markov / value-equivalence approximation error controlled by $\mathcal{L}_{\mathrm{decision}}$.\hfill$\square$

\subsection{Proof of Proposition~\ref{prop:bifurcation} and Corollary~\ref{cor:symlog}}

By the tower property, $\mathcal{B}_{\mathrm{joint}}(z,\omega) = \mathbb{E}_{Z'}[\mathbb{E}[R \mid z,Z',\omega,h] + \mathbb{E}[\tilde{\Gamma} \mid z,Z',\omega,h] V_{\bar{\xi}}(Z') \mid z,\omega,h] = \mathbb{E}[R + \tilde{\Gamma} V_{\bar{\xi}}(Z') \mid z,\omega,h] = \mathcal{B}^\star(z,\omega)$. Under ideal marginal expectations, $\mathcal{B}_{\mathrm{marg}} - \mathcal{B}^\star = \mathbb{E}[\tilde{\Gamma}]\mathbb{E}[V_{\bar{\xi}}(Z')] - \mathbb{E}[\tilde{\Gamma} V_{\bar{\xi}}(Z')] = -\mathrm{Cov}(\tilde{\Gamma}, V_{\bar{\xi}}(Z') \mid z,\omega,h)$, proving Proposition~\ref{prop:bifurcation}.

For Corollary~\ref{cor:symlog}, let $Y = \mathrm{symlog}(D) = \ln(1+D)$ for $D > 0$. Because two-hot encoding preserves exact linear expectation $\sum_i y_i(Y) b_i = Y$ on $[b_1, b_{101}] = [-20, 20]$ whenever $Y \in [-20, 20]$, minimizing cross-entropy $\mathbb{E}[-\sum_i y_i(Y) \ln p_i(z,\omega,h)]$ yields $\sum_i p_i^\star b_i = \mathbb{E}[Y \mid z,\omega,h] = \mathbb{E}[\ln(1+D) \mid z,\omega,h]$. Applying $\mathrm{symexp}(u) = e^u - 1$ for $u > 0$ gives $\hat{D}_{\mathrm{marg}} = \exp(\mathbb{E}[\ln(1+D)]) - 1 \le \mathbb{E}[D]$ by Jensen's inequality (strict whenever $\mathrm{Var}(D) > 0$). Adding and subtracting $\mathbb{E}[R] + \mathbb{E}[\gamma^D]\mathbb{E}[V_{\bar{\xi}}(Z')]$ from $\widehat{\mathcal{B}}_{\mathrm{marg,plug}} - \mathcal{B}^\star = \hat{R}_{\mathrm{marg}} + \gamma^{\hat{D}_{\mathrm{marg}}}\mathbb{E}[V_{\bar{\xi}}(Z')] - \mathbb{E}[R + \gamma^D V_{\bar{\xi}}(Z')]$ immediately yields the three terms in Eq.~\eqref{eq:three_term}. When $(D, R)$ are deterministic functions of $(z, Z', \omega, h)$, conditional expectations given $Z'$ collapse to point masses, so $\hat{D}(z,Z') = D$, $\hat{R}(z,Z') = R$, and all three bias terms vanish identically.\hfill$\square$

\subsection{Proof of Proposition~\ref{prop:ratchet} and Spherical Geodesic Transport}

For a single simplex block $p \in \Delta_{m-1}$, convex combinations $p_\tau = (1-\tau)p_0 + \tau p_1$ of distinct peaked boundary codes satisfy $\|p_\tau\|_2^2 < \max(\|p_0\|_2^2, \|p_1\|_2^2)$ and strict concavity of Shannon entropy $H(p_\tau) > (1-\tau)H(p_0) + \tau H(p_1)$, placing finite-step Euler endpoints $\hat{z}' \in \mathcal{M}_{\mathrm{gen}}$ strictly inside the higher-entropy interior. When $\alpha=0$, $\nabla_\psi \mathcal{L}_{\mathrm{emp}}$ is supported solely on $\mathcal{M}_{\mathrm{clean}}$. Taking expectation over $\hat{z}'_1, \hat{z}'_2 \in \mathcal{M}_{\mathrm{gen}}$ in $\hat{R}_{\mathrm{comp}}(2h) - R^\star(2h) = (\hat{R}_{\bar{\psi}}(z,\hat{z}'_1) - R^\star_1) + \hat{a}_1\gamma^{\hat{D}_1}(\hat{R}_{\bar{\psi}}(\hat{z}'_1,\hat{z}'_2) - R^\star_2) + (\hat{a}_1\gamma^{\hat{D}_1} - a_1^\star\gamma^{D_1^\star})R^\star_2$, under zero-mean true rewards ($R^\star \approx 0$ on $\mathcal{O}_{\mathrm{rand}}$) and homogeneous interior bias $\mathbb{E}[\hat{R}_{\bar{\psi}} - R^\star \mid \mathcal{M}_{\mathrm{gen}}] = \mu_h > 0$ independent of $\hat{a}_1\gamma^{\hat{D}_1}$, we obtain $\mu_{2h} = \mu_h + \bar{\kappa}_h \mu_h = (1 + \bar{\kappa}_h)\mu_h$. Since $\hat{a}_1 \in (0,1]$ and $\gamma^{\hat{D}_1} \in [\gamma^h, 1)$ are strictly positive, $1 + \bar{\kappa}_h > 1$ holds \emph{without} requiring $\gamma^{\hat{D}_1} \to 1$. Setting $\alpha > 0$ introduces a direct cross-entropy restoring gradient toward bounded empirical returns $R \in [-R_{\mathrm{emp}}, R_{\mathrm{emp}}]$ on $\mathcal{M}_{\mathrm{gen}}$, bounding $\mu_h$. Geometrically, on $\mathbb{S}^{d-1}(\rho) = \{z \in \mathbb{R}^d : \|z\|_2 = \rho\}$ ($\rho=8, d=64$), geodesic slerp $z_\tau = \frac{\sin((1-\tau)\Omega)}{\sin\Omega}z_0 + \frac{\sin(\tau\Omega)}{\sin\Omega}z_1$ and exponential-map Euler steps $\mathrm{Exp}_z(\Delta\tau\, v_\perp) = \cos(\frac{\Delta\tau\|v_\perp\|}{\rho})z + \sin(\frac{\Delta\tau\|v_\perp\|}{\rho})\frac{\rho v_\perp}{\|v_\perp\|}$ preserve $\|z\|_2 \equiv \rho$ identically, eliminating radial interior drift.\hfill$\square$

\section{Benchmark Environments, Skill Libraries, and Exact Hyperparameters}
\label{app:setup}

\textbf{Code-Verified Hyperparameters and Evaluation Alignment.} Table~\ref{tab:env_skills} lists physical task dimensions and skill coverage statistics. Every native benchmark and causal intervention model uses latent dimension $d=64$ ($8$ groups $\times$ $8$-way \texttt{SimNorm} with temperature $0.125$, or $\mathbb{S}^{63}(8)$ radius $\rho=8.0$), MLP hidden width $\mathbf{256}$, batch size $\mathbf{128}$ ($96$ empirical rows $+ 32$ self-distillation rows), \textbf{constant} AdamW learning rate $\mathbf{3\times 10^{-4}}$ (weight decay $10^{-4}$, no cosine decay), global gradient norm clip $\mathbf{1.0}$, EMA target decay $\mathbf{0.999}$, discount $\gamma = 0.99$, and 101-bin \texttt{Symlog-Twohot} cross-entropy on $[-20, 20]$. The detached value probe $V_\xi(\mathrm{sg}(z_t))$ is trained via \texttt{Symlog-Twohot} cross-entropy against empirical Monte Carlo discounted episode return-to-go $G_t = \sum_{k\ge 0}\gamma^k r_{t+k}$, with its EMA copy $V_{\bar{\xi}}$ providing terminal MPC bootstrap values without backpropagating into $f_\theta$. Offline \texttt{Test} splits ($200$ episodes per task-regime) are collected from 200 reserved environment reset seeds; closed-loop MPC resets the live environment to those \emph{exact same 200 initial state seeds} and interacts online until episode termination without accessing offline \texttt{Test} actions or future states.

\textbf{Exact Run Accounting.} Our controlled 50,000-update sensitivity and geometry suite executes exactly \textbf{52 new training runs}: $16$ \texttt{JSK-Flow} loss-ratio runs + $32$ non-\texttt{SimNorm} geometry runs ($4$ alternative manifolds $\times$ $8$ conditions) + $4$ analytical bifurcating-corridor runs. In unified 50k comparison tables, the $8$ \texttt{SimNorm} geometry rows alias the matching default \texttt{JSK-Flow} runs, and the $32$ \texttt{Time-Shortcut}/\texttt{Marginal-Heads} 50k rows evaluate saved 50k checkpoints from the main 100k runs with zero duplicate training ($52 + 8 + 32 - 4\text{ corridor} = 88$ tabular evaluation rows across Table~\ref{tab:geometry} and Appendix~\ref{app:ledgers}). Combined with $192$ main 100k models, $15$ \texttt{JSK-Flow-Anchored} 100k runs, $24$ spherical 100k runs, $18$ truncated $H_{\mathrm{train}} \le 16$ 100k runs, and $18$ bimodal stochastic 60k runs (plus $18$ frozen-backbone readout-only runs), the study comprises $319$ end-to-end trained world models ($337$ total training runs).

\begin{labtable}[!htb]{Environment physical specifications and held-out validation option statistics ($H=128$, $|\Omega|=8$ options per library; 2,000 / 200 / 200 Train / Val / Test splits). ``Pos.\ Rate'' denotes episodes triggering $>0$ reward; ``1-Opt.\ Solve'' denotes full episode completion by a single option rollout. \texttt{AntMaze} uses \texttt{AntMaze\_UMazeDense-v5} (\texttt{mujoco 3.17.0}).}{tab:env_skills}
{>{\hsize=0.90\hsize\linewidth=\hsize\raggedright\arraybackslash}X
 >{\hsize=1.22\hsize\linewidth=\hsize\centering\arraybackslash}X
 >{\hsize=0.76\hsize\linewidth=\hsize\centering\arraybackslash}X
 >{\hsize=0.96\hsize\linewidth=\hsize\centering\arraybackslash}X
 >{\hsize=1.04\hsize\linewidth=\hsize\centering\arraybackslash}X
 >{\hsize=1.12\hsize\linewidth=\hsize\centering\arraybackslash}X}
\labh{Task} & \labh{Obs / Act / $H_{\max}$} & \labh{$\sigma_{\mathrm{ret}}$} & \labh{$\mathcal{O}_{\mathrm{rand}}$ Span} & \labh{$\mathcal{O}_{\mathrm{skill}}$ Span} & \labh{Pos.\ / 1-Opt.\ Solve}\\\labheadrule
FourRooms & $208$ / $4$ disc.\ / $200$ & 1.000 & 19.8 steps & 32.6 steps & 37.0\% / 37.0\%\\\labrowrule
PinPad & $71$ / $5$ disc.\ / $1000$ & 1.000 & 20.4 steps & 60.9 steps & 24.0\% / 22.8\%\\\labrowrule
Sokoban & $700$ / $4$ disc.\ / $200$ & 4.361 & 17.4 steps & 51.9 steps & 48.0\% / 3.0\%\\\labrowrule
AntMaze & $109$ / $8$ cont.\ / $700$ & 1.000 & 18.9 steps & 63.4 steps & \textemdash\ / 6.0\% (reach)\\
\end{labtable}

\section{Complete Compute, Execution-Mode, Causal Swap, and Sensitivity Ledgers}
\label{app:ledgers}

Table~\ref{tab:exec_costs} reports the paired closed-loop solve rates and compute compression ratios comparing primitive replanning against option-to-termination execution. Table~\ref{tab:full_budget_grid} reports the complete six-budget closed-loop MPC grid across $\mathcal{O}_{\mathrm{rand}}$, $\mathcal{O}_{\mathrm{skill}}$, and combined evaluation scopes. Table~\ref{tab:causal_mechanisms} details the zero-retraining single-head checkpoint swaps and frozen-backbone readout isolation runs, and Table~\ref{tab:per_task_pred} reports the 50,000-update sensitivity summary.

\begin{longtable}{@{\hspace{3pt}}P{0.22}Q{0.11}Q{0.11}Q{0.12}Q{0.18}Q{0.13}Q{0.13}@{\hspace{3pt}}}
\caption{Primitive replanning versus option-to-termination execution at the default budget ($B=8, C=64, N_p=4$). Solve rates (\%) and paired gain $\Delta = \text{Opt} - \text{Prim}$ (percentage points) are reported on $\mathcal{O}_{\mathrm{skill}}$ (100k checkpoints); PNFE ratios report both pooled total ratio ($\sum\mathrm{PNFE}_{\mathrm{prim}}/\sum\mathrm{PNFE}_{\mathrm{opt}}$) and 144-cell median ratio alongside exact per-decision row queries and wall-clock latency.}
\label{tab:exec_costs}\\
\labtoprule
\labh{Model Family} & \labh{Prim.\ \%} & \labh{Opt.\ \%} & \labh{Paired $\Delta$} & \labh{Pooled / Med.\ $\times$} & \labh{Queries} & \labh{ms/Dec}\\\labheadrule
\endfirsthead
\labtoprule
\labh{Model Family} & \labh{Prim.\ \%} & \labh{Opt.\ \%} & \labh{Paired $\Delta$} & \labh{Pooled / Med.\ $\times$} & \labh{Queries} & \labh{ms/Dec}\\\labheadrule
\endhead
\endfoot
\endlastfoot
Expectation-Option & \textbf{53.62} & \textbf{55.17} & +1.54 & $27.2\times$ / $21.0\times$ & 4,609 & 26.75\\\labrowrule
n-Step-MC & 51.79 & 54.71 & +2.92 & $27.0\times$ / $19.6\times$ & 3,585 & \textbf{24.39}\\\labrowrule
Denoise-Only & 40.21 & 51.79 & +11.58 & $25.5\times$ / $18.1\times$ & 3,585 & 25.36\\\labrowrule
Geometric-TD-HC & 46.33 & 49.29 & +2.96 & $26.2\times$ / $19.5\times$ & 3,585 & 28.83\\\labrowrule
Time-Shortcut & 26.77 & 45.09 & \textbf{+18.32} & $26.1\times$ / $21.1\times$ & 4,609 & 31.36\\\labrowrule
One-Step & 31.75 & 40.96 & +9.21 & $26.4\times$ / \textbf{25.1$\times$} & 6,657 & 31.81\\\labrowrule
JSK-Flow (unanc.) & 34.12 & 40.92 & +6.79 & $26.2\times$ / $20.4\times$ & 3,585 & 24.42\\\labrowrule
Marginal-Heads & 31.67 & 35.12 & +3.46 & $26.7\times$ / $19.9\times$ & 3,585 & 33.36\\\labrowrule
\end{longtable}

\begin{longtable}{@{\hspace{3pt}}P{0.25}Q{0.125}Q{0.125}Q{0.125}Q{0.125}Q{0.125}Q{0.125}@{\hspace{3pt}}}
\caption{Complete closed-loop Test solve rates (\%) across all six compute budgets ($C=64$; column headers list $(B, N_p)$ and Dynamics Transition PNFE per decision: $256, 1024, 512, 2048, 1024, 4096$; 439,200 total terminal checkpoint episodes). \textemdash\ denotes budget-infeasible cells ($K_{\mathrm{jump}}=8 > B=4$).}
\label{tab:full_budget_grid}\\
\labtoprule
\labh{Scope / Family} & \labh{B4 Np1 (256)} & \labh{B4 Np4 (1024)} & \labh{B8 Np1 (512)} & \labh{B8 Np4 (2048)} & \labh{B16 Np1 (1024)} & \labh{B16 Np4 (4096)}\\\labheadrule
\endfirsthead
\labtoprule
\labh{Scope / Family} & \labh{B4 Np1 (256)} & \labh{B4 Np4 (1024)} & \labh{B8 Np1 (512)} & \labh{B8 Np4 (2048)} & \labh{B16 Np1 (1024)} & \labh{B16 Np4 (4096)}\\\labheadrule
\endhead
\endfoot
\endlastfoot
$\mathcal{O}_{\mathrm{rand}}$: Expectation-Option & \textbf{10.79} & \textbf{10.77} & \textbf{10.75} & \textbf{10.67} & \textbf{10.31} & \textbf{10.94}\\\labrowrule
$\mathcal{O}_{\mathrm{rand}}$: n-Step-MC & 9.52 & 10.29 & 10.19 & 10.15 & 9.50 & 10.00\\\labrowrule
$\mathcal{O}_{\mathrm{rand}}$: Denoise-Only & \textemdash & \textemdash & 9.15 & 8.52 & 8.52 & 8.98\\\labrowrule
$\mathcal{O}_{\mathrm{rand}}$: Geometric-TD-HC & 8.10 & 8.27 & 7.96 & 8.40 & 7.77 & 8.19\\\labrowrule
$\mathcal{O}_{\mathrm{rand}}$: Time-Shortcut & 7.94 & 8.35 & 7.75 & 8.35 & 7.96 & 8.69\\\labrowrule
$\mathcal{O}_{\mathrm{rand}}$: Marginal-Heads & 7.79 & 7.44 & 8.38 & 7.56 & 7.81 & 8.00\\\labrowrule
$\mathcal{O}_{\mathrm{rand}}$: One-Step & 7.65 & 8.00 & 7.44 & 7.46 & 8.00 & 7.73\\\labrowrule
$\mathcal{O}_{\mathrm{rand}}$: JSK-Flow & 7.15 & 6.21 & 7.25 & 7.31 & 7.42 & 7.40\\\labrowrule
$\mathcal{O}_{\mathrm{skill}}$: Expectation-Option & \textbf{54.75} & \textbf{55.04} & \textbf{54.29} & \textbf{54.40} & \textbf{54.21} & \textbf{54.27}\\\labrowrule
$\mathcal{O}_{\mathrm{skill}}$: n-Step-MC & 52.98 & 53.77 & 52.23 & 53.25 & 52.23 & 53.29\\\labrowrule
$\mathcal{O}_{\mathrm{skill}}$: Geometric-TD-HC & 43.88 & 46.52 & 42.90 & 47.81 & 45.08 & 48.12\\\labrowrule
$\mathcal{O}_{\mathrm{skill}}$: Denoise-Only & \textemdash & \textemdash & 47.54 & 46.00 & 46.52 & 47.23\\\labrowrule
$\mathcal{O}_{\mathrm{skill}}$: JSK-Flow & 37.27 & 37.38 & 37.52 & 37.52 & 36.54 & 36.96\\\labrowrule
$\mathcal{O}_{\mathrm{skill}}$: One-Step & 37.35 & 37.46 & 36.23 & 36.35 & 36.50 & 36.54\\\labrowrule
$\mathcal{O}_{\mathrm{skill}}$: Time-Shortcut & 36.52 & 35.61 & 36.75 & 35.93 & 35.95 & 35.27\\\labrowrule
$\mathcal{O}_{\mathrm{skill}}$: Marginal-Heads & 28.62 & 30.40 & 32.04 & 33.40 & 32.83 & 34.33\\\labrowrule
Combined: Expectation-Option & \textbf{32.77} & \textbf{32.91} & \textbf{32.52} & \textbf{32.53} & \textbf{32.26} & \textbf{32.60}\\\labrowrule
Combined: n-Step-MC & 31.25 & 32.03 & 31.21 & 31.70 & 30.86 & 31.65\\\labrowrule
Combined: Geometric-TD-HC & 25.99 & 27.40 & 25.43 & 28.10 & 26.43 & 28.16\\\labrowrule
Combined: Denoise-Only & \textemdash & \textemdash & 28.34 & 27.26 & 27.52 & 28.10\\\labrowrule
Combined: JSK-Flow & 22.21 & 21.79 & 22.39 & 22.42 & 21.98 & 22.18\\\labrowrule
Combined: One-Step & 22.50 & 22.73 & 21.83 & 21.91 & 22.25 & 22.14\\\labrowrule
Combined: Time-Shortcut & 21.61 & 21.39 & 21.62 & 21.54 & 21.35 & 21.40\\\labrowrule
Combined: Marginal-Heads & 18.21 & 18.92 & 20.21 & 20.48 & 20.32 & 21.17\\\labrowrule
\end{longtable}

\begin{longtable}{@{\hspace{3pt}}P{0.30}Q{0.14}Q{0.14}Q{0.14}Q{0.14}Q{0.14}@{\hspace{3pt}}}
\caption{Zero-retraining single-head checkpoint swaps ($H=128$ stopped-option mode on \texttt{AntMaze}) and frozen-backbone Joint vs.\ Marginal readout isolation ($\Delta\mathrm{StateMSE} \equiv 0$, 3 seeds per task on $\mathcal{O}_{\mathrm{skill}}$).}
\label{tab:causal_mechanisms}\\
\labtoprule
\labh{Intervention / Variant} & \labh{Scope} & \labh{Return MAE} & \labh{Value Error} & \labh{Span MAE} & \labh{Opt.\ MPC \%}\\\labheadrule
\endfirsthead
\labtoprule
\labh{Intervention / Variant} & \labh{Scope} & \labh{Return MAE} & \labh{Value Error} & \labh{Span MAE} & \labh{Opt.\ MPC \%}\\\labheadrule
\endhead
\endfoot
\endlastfoot
Full 100k (saturated) & AntMaze $\mathcal{O}_{\mathrm{rand}}$ s1 & $4.852\mathrm{e}8$ & $4.852\mathrm{e}8$ & 108.31 & 1.50\\\labrowrule
Gen100k + Swap $(D,m,e)$ 10k & AntMaze $\mathcal{O}_{\mathrm{rand}}$ s1 & $4.852\mathrm{e}8$ & $4.852\mathrm{e}8$ & \textbf{15.28} & 0.50\\\labrowrule
Gen100k + Swap $R$-Only 10k & AntMaze $\mathcal{O}_{\mathrm{rand}}$ s1 & \textbf{0.129} & 0.699 & 108.31 & 1.00\\\labrowrule
Gen100k + Swap All 10k & AntMaze $\mathcal{O}_{\mathrm{rand}}$ s1 & \textbf{0.129} & \textbf{0.177} & \textbf{15.28} & 0.50\\\labrowrule
Full 100k (late drift) & AntMaze $\mathcal{O}_{\mathrm{skill}}$ s2 & 57.60 & 57.58 & 67.35 & 9.50\\\labrowrule
Gen100k + Swap $R$-Only 10k & AntMaze $\mathcal{O}_{\mathrm{skill}}$ s2 & \textbf{0.577} & \textbf{0.713} & 67.35 & \textbf{21.50}\\\labrowrule
Frozen Backbone: Joint Head & FourRooms $\mathcal{O}_{\mathrm{skill}}$ & \textbf{0.249} & 0.181 & \textbf{20.38} & 99.17\\\labrowrule
Frozen Backbone: Marginal Head & FourRooms $\mathcal{O}_{\mathrm{skill}}$ & 0.252 & 0.181 & 20.47 & \textbf{99.50}\\\labrowrule
Frozen Backbone: Joint Head & PinPad $\mathcal{O}_{\mathrm{skill}}$ & \textbf{1.891} & 1.907 & \textbf{40.95} & 49.17\\\labrowrule
Frozen Backbone: Marginal Head & PinPad $\mathcal{O}_{\mathrm{skill}}$ & 1.893 & 1.907 & 41.52 & \textbf{52.67}\\\labrowrule
Frozen Backbone: Joint Head & AntMaze $\mathcal{O}_{\mathrm{skill}}$ & \textbf{0.359} & \textbf{0.545} & 38.57 & \textbf{17.17}\\\labrowrule
Frozen Backbone: Marginal Head & AntMaze $\mathcal{O}_{\mathrm{skill}}$ & 0.361 & 0.558 & \textbf{38.55} & 16.33\\\labrowrule
\end{longtable}

\begin{longtable}{@{\hspace{3pt}}P{0.28}Q{0.18}Q{0.18}Q{0.18}Q{0.18}@{\hspace{3pt}}}
\caption{Controlled 50,000-update sensitivity summary across the 16-condition native grid (4 tasks $\times$ 2 regimes $\times$ 2 seeds; $H=128$ stopped-option Test evaluation on \texttt{SimNorm}).}
\label{tab:per_task_pred}\\
\labtoprule
\labh{Model Family (50k)} & \labh{Med.\ Return MAE} & \labh{Max Return MAE} & \labh{Med.\ Value Error} & \labh{Med.\ Span MAE}\\\labheadrule
\endfirsthead
\labtoprule
\labh{Model Family (50k)} & \labh{Med.\ Return MAE} & \labh{Max Return MAE} & \labh{Med.\ Value Error} & \labh{Med.\ Span MAE}\\\labheadrule
\endhead
\endfoot
\endlastfoot
Time-Shortcut (50k) & 0.273 & 1.894 & \textbf{0.278} & 17.41\\\labrowrule
Marginal-Heads (50k) & \textbf{0.237} & \textbf{1.848} & 2.011 & \textbf{16.78}\\\labrowrule
JSK-Flow (50k, unanchored) & 0.675 & $1.882\mathrm{e}7$ & 6.496 & 31.19\\\labrowrule
\end{longtable}

\end{document}